\documentclass[10pt,twocolumn]{article} 
\usepackage{simpleConference}
\usepackage{times}
\usepackage{graphicx}
\usepackage{amssymb}
\usepackage{url,hyperref}

\usepackage{amssymb}
\usepackage{color}
\usepackage{array}
\usepackage{multirow}
\usepackage{arydshln}
\usepackage{booktabs}
\usepackage{multirow}
\usepackage{graphicx}
\usepackage{arydshln}
\usepackage{amsmath}
\usepackage{mathtools}
\usepackage{stackengine}
\usepackage{marvosym}
\usepackage[numbers]{natbib}
\usepackage[font=small,labelfont=bf,labelsep=colon]{caption}
\usepackage{hyperref}
\begin{document}


\title{\large LayerEdit: Disentangled Multi-Object Editing via Conflict-Aware Multi-Layer Learning}

\author{
Fengyi Fu\textsuperscript{\rm 1}, Mengqi Huang\textsuperscript{\rm 1  \Letter} \thanks{\Letter \ Mengqi Huang is the corresponding author}, Lei Zhang\textsuperscript{\rm 1}, Zhendong Mao\textsuperscript{\rm 1} 
\\
\textsuperscript{\rm 1}University of Science and Technology of China,\\
\{ff142536f, huangmq\}@mail.ustc.edu.cn, \{leizh23, zdmao\}@ustc.edu.cn
}

\maketitle
\thispagestyle{empty}

\begin{abstract}
Text-driven multi-object image editing which aims to precisely modify multiple objects within an image based on text descriptions, has recently attracted considerable interest. Existing works primarily follow the {localize-editing} paradigm, 
focusing on independent object localization and editing while neglecting critical inter-object interactions.
However, this work points out that the neglected attention entanglements in inter-object conflict regions, inherently hinder disentangled multi-object editing, leading to either inter-object editing leakage or intra-object editing constraints.
We thereby propose a novel multi-layer disentangled editing framework \textbf{\textit{LayerEdit}}, a training-free method which, for the first time, through precise object-layered decomposition and coherent fusion, enables conflict-free object-layered editing.
Specifically, \textit{LayerEdit} introduces a novel ``decompose-editing-fusion" framework, consisting of:
(1) \textit{Conflict-aware Layer Decomposition module}, which utilizes an attention-aware IoU scheme and time-dependent region removing, to enhance conflict awareness and suppression for layer decomposition. 
(2) \textit{Object-layered Editing module}, to establish coordinated intra-layer text guidance and cross-layer geometric mapping, achieving disentangled semantic and structural modifications.
(3) \textit{Transparency-guided Layer Fusion module}, to facilitate structure-coherent inter-object layer fusion through precise transparency guidance learning.
Extensive experiments verify the superiority of \textit{LayerEdit} over existing methods, showing unprecedented intra-object controllability and inter-object coherence in complex multi-object scenarios. Codes are available at:
\href{https://github.com/fufy1024/LayerEdit}{https://github.com/fufy1024/LayerEdit}.

\end{abstract}

\begin{figure}[t]
\centering
\includegraphics[width=0.95\columnwidth]{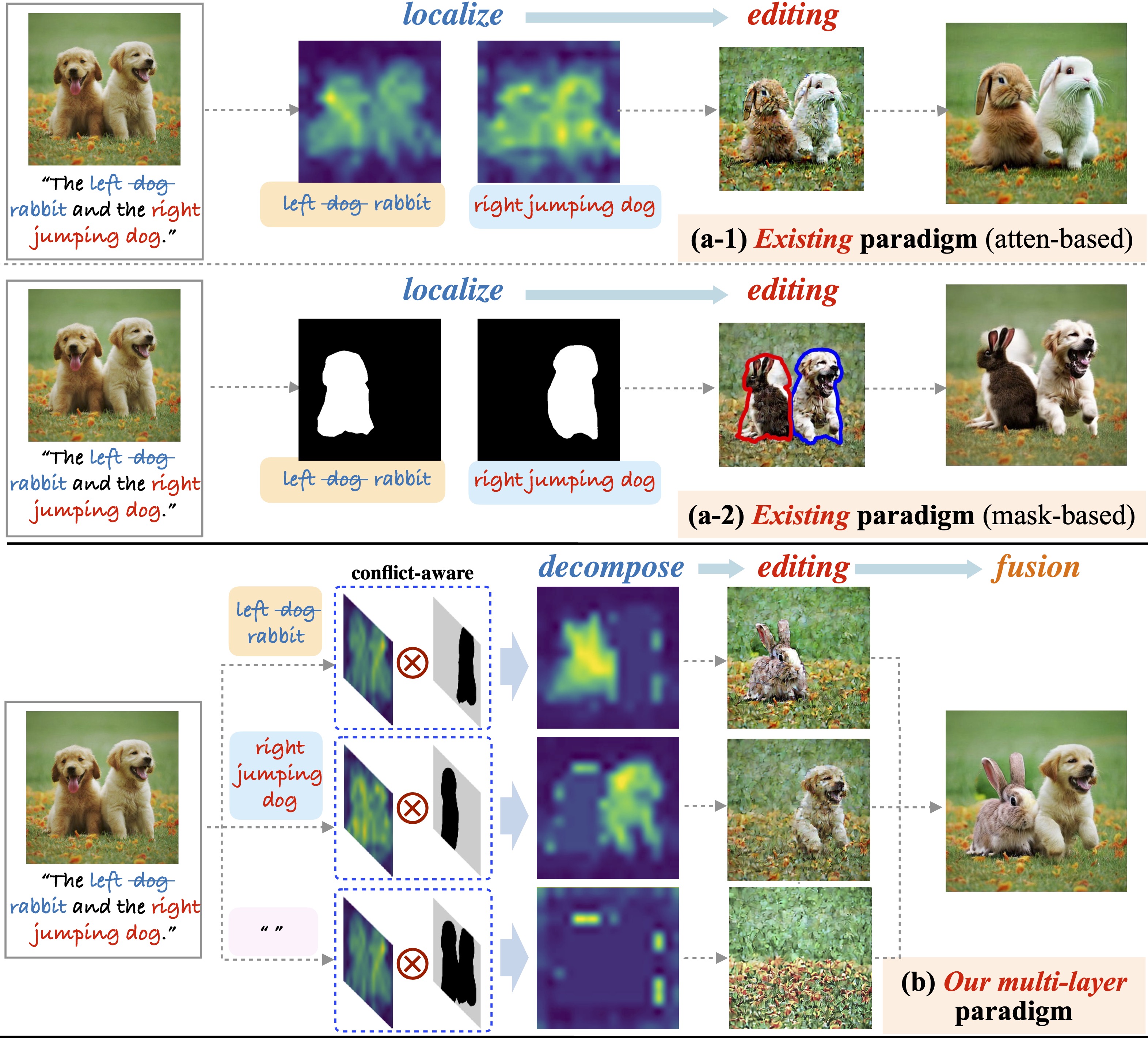} 
\caption{
{Illustration of motivation. (a) \textit{Existing paradigm}}: suffers from inaccurate disentanglement across conflict regions, resulting in (a-1) inter-object editing leakage or (a-2) intra-object editing artifacts. {(b) \textit{Our multi-layer disentangled editing paradigm}}: by modeling conflict-aware object-layered decomposition and structure-coherent fusion,
bringing conflict-free and accurate multi-object editing.
}
\label{fig1}
\end{figure}

\section{Introduction}

Recent diffusion-based image generation models have revolutionized the Text-driven Image Editing (TIE) task \cite{wei2025omniedit}. 
TIE aims to perform arbitrary text-driven image modifications with a good preservation of irrelevant regions, offering powerful applications for advertising, photography, social media, and so on. 
Given that real-world images typically consist of multiple objects with intricate spatial and attribute relationships, recent methods \cite{chakrabarty2024lomoe} extend TIE to more complex text-driven multi-object editing (TMOE) scenarios.  
Compared to simple single-object editing, the primary goal of TMOE is dual-faceted:
(1) \textbf{\textit{intra-object controllability}}, \textit{i.e.}, 
precise text-controlled editing for each target object, 
without being constrained by its original spatial boundaries;
(2) \textbf{\textit{inter-object coherence}}, \textit{i.e.}, 
coherent object-specific editing across multiple objects, 
without unintended modifications on non-target objects.

Recent works take their effort on two aspects toward these goals. 
The first relies on attention-based text-image alignment \cite{epstein2024diffusion,guo2024focus,huang_etal_neurips24c} for target object localization, by iterative object-specific editing at multi-step \cite{huang_etal_neurips24c} or multi-turn \cite{brack2024ledits++} to maintain multi-object coherence.
The other methods incorporate segmentation \cite{chakrabarty2024lomoe,yang2024object} or visual language models \cite{wang2024genartist,feng2024ranni,schouten2025poem} to obtain more accurate object location, combined with loss regularization \cite{zhu2025mde} or mask-weighted fusion \cite{chakrabarty2024lomoe,yang2024object} to minimize unintended modifications in non-target regions.
In general, existing methods primarily follow a \textit{localize-editing} paradigm, \textit{i.e.}, 
independently performing intra-object localizing and editing, 
without considering inter-object interactions.

\begin{figure}[t]
\centering\includegraphics[width=0.93\columnwidth]{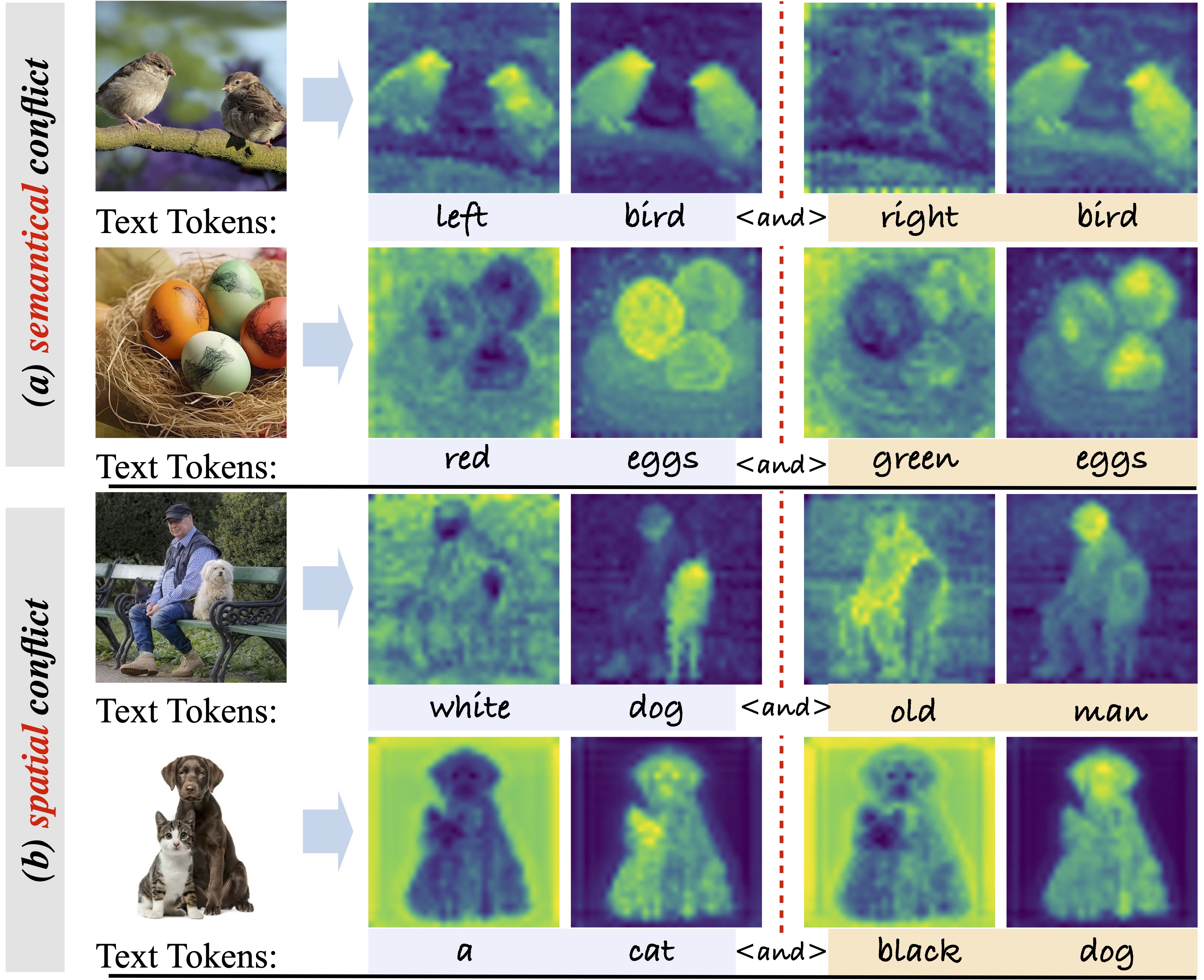} 
\caption{{Visualization of cross-attention maps} obtained from SDXL, associated with different tokens. 
The key observation is that model exhibits text-image attention misalignment in both: {(a)} \textit{semantical}, and {(b)} \textit{spatial conflict regions}.
}
\label{figattn}
\end{figure}


However, we argue that existing frameworks, fixated on intra-object editing while ignoring inter-object interactions, inevitably hinder accurate disentanglement across multiple objects, ultimately compromising either inter-object coherence or intra-object controllability.
The fundamental limitation stems from text-conditioned diffusion models' reliance on text-image attention alignment for feature manipulation: while effective for single-object editing, they catastrophically fail in multi-object scenarios due to attention entanglement caused by unmodeled inter-object interactions.
 As analyzed in Fig.\textcolor{blue}{\ref{figattn}}, initial denoising cross-attention maps 
 reveal two characteristic attention alignment conflict patterns: 
 (1) semantic conflicts in conceptually related regions (\textit{e.g.}, Fig.\textcolor{blue}{\ref{figattn}.a}, multi-directional ``birds" or multi-color ``eggs"), and  (2) spatial conflicts in overlapping regions (\textit{e.g.}, Fig.\textcolor{blue}{\ref{figattn}.b}, occluded ``cat" and ``dog").  
 Such regions exhibiting semantic/spatial conflicts with target regions constitute \textbf{\textit{conflict regions}}, fundamentally hindering object disentanglement, 
  posing limitations to two core editing goals: 
  (1) Attention-based methods suffer from inadequate attention disentanglement of inter-object conflicts, causing inter-object editing leakage (\textit{e.g.}, Fig.\textcolor{blue}{\ref{fig1}.a-1} misediting "right dog" to "rabbit"). 
   (2) Mask-based methods suffer from non-adaptive disentanglement with fixed intra-object spatial constraints, suppressing intra-object editing in necessary structural changes (\textit{e.g.}, Fig.\textcolor{blue}{\ref{fig1}.a-2}, artifacts during "rabbit" morphing). 
Therefore, a reasonable TMOE framework should establish comprehensive multi-object disentanglement, ensuring both precise and region-unconstrained intra-object disentanglement, and conflict-constrained inter-object disentanglement.



To this end, we present \textbf{\textit{LayerEdit}}, a novel and training-free framework that reformulates multi-object editing as a multi-layer disentangled learning problem, 
which for the first time to our knowledge, explicitly models object-layered decomposition and coherent fusion, to achieve disentangled object editing free from inter-object conflicts and intra-object region constraints (Fig.\textcolor{blue}{\ref{fig1}.b}). 
Departing from conventional paradigm, the \textit{core idea of} \textit{LayerEdit} lies in fundamentally shifting the focus from mere intra-object target regions to inter-object conflict regions, which enables: conflict-free object disentangled editing through inter-object conflict awareness and suppression;
and structure-coherent object-layered fusion through inter-object structural modeling.
Technically, {\textit{LayerEdit}} introduces an innovative ``decompose-editing-fusion" architecture,  consisting of:
(1) \textit{Conflict-aware Layer Decomposition module}, which integrates cross-attention and panoptic segmentations via an adaptive IoU (Intersection over Union) scheme for accurate conflict region identification, followed by a time-dependent region removing scheme for conflict-free layer decomposition.  
(2) \textit{Object-layered Editing module}, which establishes coordinated intra-layer text guidance and cross-layer geometric mapping, for disentangled semantic and structural modifications. 
(3) \textit{Transparency-guided Layer Fusion module}, which implements layer transparency learning conditioned on inter-object spatial features, to ensure structure-coherent fusion of overlapping multi-layer.

Our contributions are summarized as follows: (1) For the first time, we point out that existing frameworks ignore inter-object interaction conflicts, resulting in either inter-object editing leakage or intra-object editing constraint. We propose \textit{LayerEdit}, a \textit{training-free} and \textit{plug-and-play} framework that, through precise object-layered decomposition and coherent multi-layer fusion, achieves disentangled object editing free from inter-object conflicts and intra-object constraints.
(2) We propose a novel decompose-editing-fusion framework, with three tailored modules to respectively enhance the conflict-free object layer decomposition, the disentangled editing for arbitrary semantic and structural modifications, and the coherent fusion for overlapping layers.
(3) Extensive experiments validate \textit{LayerEdit}'s superiority, demonstrating significant improvements in both editability ($12.6\%$ in CLIP-T) and  global quality ($15.3\%$ in FID). 


\begin{figure*}[!t]
\centering
\includegraphics[width=0.97\linewidth]{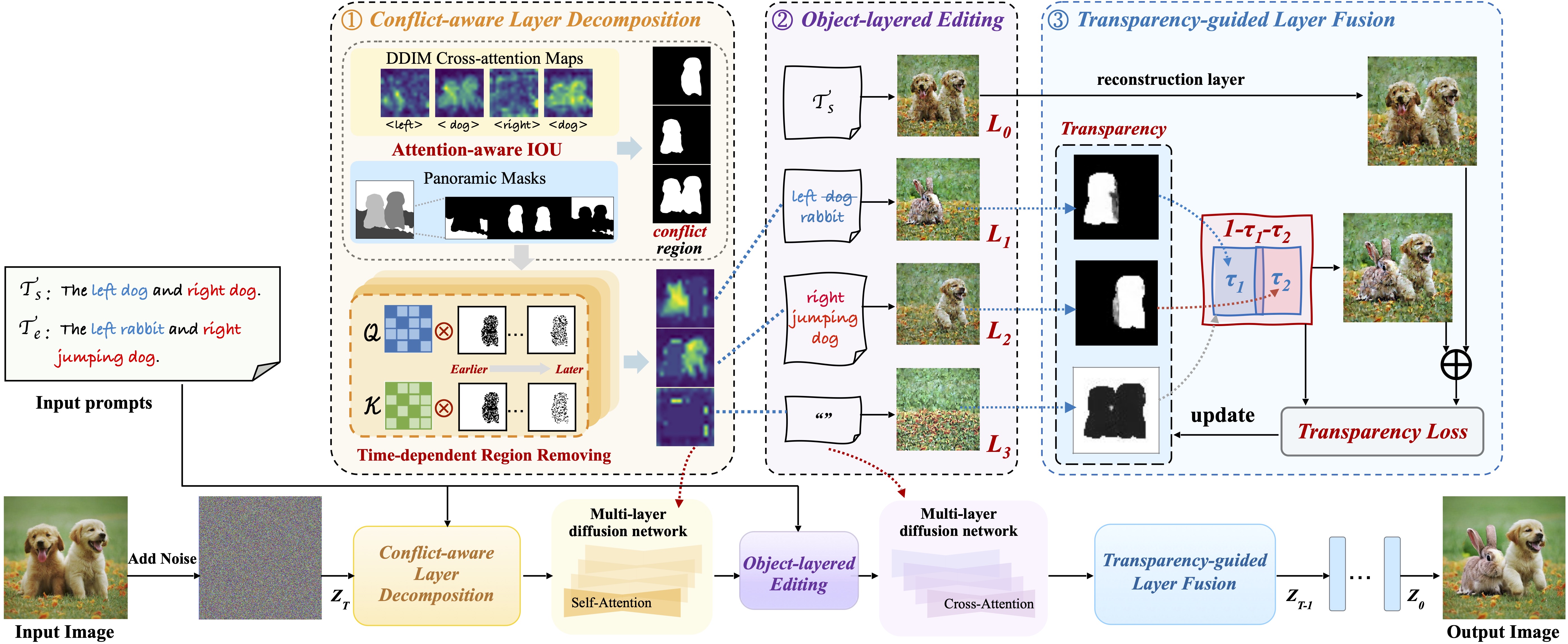} 
\caption{{Overview of \textit{LayerEdit}}, 
consisting of: 1) Conflict-aware Layer Decomposition: precisely decompose object layers by identifying and constraining conflict regions; 2) Object-layered Editing: establish intra-layer text-guided editing (geometric editing detailed in Fig.\textcolor{blue}{{\ref{ge-method}}});
 3) Transparency-guided Layer Fusion: enables structure-coherent fusion with transparency learning.
}
\label{fremework}
\end{figure*}

\section{Related Work}

\subsection{Text-Driven Image Editing}
Recent diffusion models \cite{fu2025feededit,zhou2025fireedit} revolutionize the text-driven image editing task. 
\cite{hertz2023prompt,tumanyan2023plug} first achieve the image editing through text interaction directly. 
\cite{kawar2023imagic,li2023layerdiffusion,song2024doubly} enable non-rigid image editing with complex text embedding optimization. 
\cite{brooks2023instructpix2pix, chen2025instruct} expand the task into instruction-guided editing scenarios. 
Other works inject attention-level \cite{cao2023masactrl,fu2025feededit} or task-specific \cite{epstein2024diffusion,nguyen2025h} control into generation process, facilitating precise image manipulations. 
Although effective in single-object editing, these models struggle with real-world multi-object scenarios, which have complex spatial and semantic inter-object relationships.


\subsection{Multi-Object Image Editing}
Following single-object editing work, \cite{joseph2024iterative,brack2024ledits++} first propose an iterative editing framework to ensure multi-object coherent editing in a multi-turn editing process, but result in exponentially increasing calculation and time costs.
\cite{jia2024designedit,yang2024object,chakrabarty2024lomoe} adopt segmentation models to identify objects, based on mask-weighted fusion to concatenate various edited objects and unedited backgrounds, which impose strict region constraints and struggle with layout modifications. 
\cite{patashnik2023localizing,guo2024focus,huang_etal_neurips24c,li2025moedit} implicitly distinguish objects based on cross-attention maps, but easily entangle features of spatially overlapping or semantically related objects. 
However, different from existing methods, the core idea of \textit{LayerEdit} is to explicitly model and resolve inter-object conflicts through a multi-layer decompose-editing-fusion framework, which is completely unexplored.

\section{Preliminaries}
 \label{Preliminaries}

Diffusion models consist of a forward process and a corresponding reverse process. 
The forward process gradually adds Gaussian noise to data, to generate the noisy sample with a predefined noise adding schedule $\alpha_t$ at timestep $t$:
\begin{equation}
z_t = \sqrt{{\overline{\alpha}}_t } z_0 +\sqrt{1-{\overline{\alpha}}_t } \epsilon, \ \ \  \epsilon \in \mathcal{N}(0,1),
\end{equation}
where ${\overline{\alpha}}_t = \prod^t_{i=1} \alpha_i$, $z_0$ is a sample from data distribution.

The reverse process involves a parameterized noise prediction network $\epsilon_\theta$, to predict and iteratively remove the added noise. The training objective is to minimize the distance between two noises, based on the text condition $\mathcal{T}$:
\begin{equation}
\mathcal{L}= \mathbb{E}_{{z}_0, \mathcal{T},  \epsilon \sim \mathcal{N}(0,1), t } [ \parallel \epsilon - \epsilon_\theta({z}_t, t, \mathcal{T}) \parallel]. 
\end{equation}

The feature incorporation in diffusion models is mainly based on a self-attention and cross-attention module, to update the spatial features ${\phi({z}}_t)$ as $\hat{\phi}({z}_t)$ :
\begin{align}
 \hat{\phi}({z}_t,\mathcal{T} )=  \text{Softmax}(\frac{\mathbf{\mathcal{{Q}}} \mathbf{\mathcal{{K}}} ^\top}{\sqrt d})\mathbf{\mathcal{{V}}} = \mathbf{\mathcal{{A}}}\mathbf{\mathcal{{V}}} , 
 \label{Eq-gen}
\end{align}
where $\mathcal{Q}$ is the query features projected from spatial features ${\phi({z}}_t)$, and $\mathcal{K}$, $\mathcal{V}$ are the key and value features projected from the spatial features (in self-attention layers) or textual embedding (in cross-attention layers) with corresponding projection matrices. $\mathcal{A}$ is the calculated attention maps.

\section{Method}

\noindent
\textbf{Definition.} Given a source image $\mathcal{I}$ and source text $\mathcal{T}_{s}$, the image editing task aims to synthesize the desired image with arbitrary rigid and non-rigid changes, based on editing prompt $\mathcal{T}_{e}$ (modified from $\mathcal{T}_{s}$). 
For multi-object editing scenarios, $ \langle \mathcal{T}_{s}, \mathcal{T}_e  \rangle$ can generally be represented as a unit of multiple editing pairs 
$\langle  \mathcal{T}_{s}=\bigcup_i\mathcal{O}^i_{s}, \mathcal{T}_e =\bigcup_i \mathcal{O}^i_{e}\rangle_{i=1}^N$ 
, where $\mathcal{O}^i_{s}$ and $\mathcal{O}^i_{e}$ indicate the $i$-th source-edited object in $\mathcal{T}_{s}$ and $\mathcal{T}_{e}$, and $N$ is the total number of objects to be edited.

In this section, we detail the implementation of \textit{LayerEdit}, an elegant decompose-editing-fusion framework shown in Fig.\textcolor{blue}{\ref{fremework}},
which first precisely decompose each object layer through {intra-object conflict} awareness and constraining (Sec.\textcolor{blue}{{\ref{Decomposition}}}), then explores the intra- and cross-layer feature processing to support semantic and structural editing functions (Sec.\textcolor{blue}{{\ref{editing}}}), and finally realizes coherent layer fusion with structure-conditioned transparency learning (Sec.\textcolor{blue}{{\ref{fusion}}}).

\subsection{Conflict-aware Layer Decomposition} 
\label{Decomposition}

Object-layered decomposition aims to construct exclusive editing layer for each object, free from attention entanglement and inter-object conflicts.
By analyzing the attention entanglement pattern in Fig.\textcolor{blue}{\ref{figattn}}, it stems from cross-attention misalignment in semantic or spatial conflict regions, hindering accurate localized editing.
However, in contrast, this phenomenon also reveals that cross-attention facilitates conflict regions awareness. Thus, we design a novel attention-aware IoU scheme, by integrating cross-attention maps and panoptic segmentation to precisely localize conflict regions. 

\noindent
\textbf{Attention-aware IoU.} 
Specifically, following established practices \cite{guo2024focus}, we first aggregate the average cross-attention maps  $\{\mathcal{\overline{A}}^i\}_{i=1}^N$ on object tokens $\{\mathcal{O}^i_{s}\}_{i=1}^N$ across all DDIM inversion timesteps.
Then, by using segmentation models (SAM, \cite{kirillov2023segment}; OneFormer, \cite{jain2023oneformer}), more precise object masks $\{\mathcal{M}^i_{o}\}_{i=1}^N$ and $K$ panoramic segmentation masks  $\{\mathcal{M}^j_{pan}\}_{i=1}^K$ are obtained. 
The attention-aware IoU is defined as: 
\begin{equation}
\begin{aligned}
\text{A-IoU}(i, j)  =  \frac{|| \mathcal{M}^j_{pan}  \bigcap \mathcal{\overline{A}}^i||}{|| \mathcal{M}^j_{pan}  \bigcup \mathcal{\overline{A}}^i||},\forall i \in [1,N],j\in [1,K]
\end{aligned}
 \end{equation}
Intuitively, $\text{A-IoU}(i, j)$ quantifies the affect weights of $j$-th panoptic region on the generation of $i$-th object.
The conflict region for $i$-th object is defined as the high-affect regions (determined by threshold $\eta$) excluding object’s own region:
\begin{equation}
\begin{aligned}
\mathcal{M}^i_{con}  =  \bigcup{}_{\{ j|\text{A-IoU}(i,j)> \eta \} }(\mathcal{M}^j_{pan}) - \mathcal{M}^i_{o}.
\end{aligned}
 \end{equation}

\noindent
\textbf{Time-dependent Region Removing. } 
Having identified the conflict regions, decomposing the $i$-th object equates to suppressing its corresponding conflict regions.
The \textit{core challenge} lies in removing the conflict region without disrupting the global harmonious generation.
Thus, we further design a time-dependent region removing strategy that dynamically suppresses conflict regions in self-attention features.

Specifically, we define a monotonically decreasing function $r(t)$ that varies with the image's Signal-to-Noise Ratio (SNR) at each timestep,  
along with a corresponding region feature weighting removal strategy on arbitrary feature $\mathcal{F}$: 
\begin{align}
r(t) &= \text{Sigmoid}  (k  (\frac{ \text{SNR}(t_{thres})}{ \text{SNR}(t)} -1)), \\
\text{Re}(\mathcal{F},\mathcal{M}_{con}) &=  \mathcal{F}  \odot (1- \mathcal{M}_{con} \odot \text{Bernoulli}(r)),
\end{align}
where $\text{SNR}(t)= \sqrt{\frac{    {\overline{\alpha}}_t }{ 1- {\overline{\alpha}}_t  } }$ is obtained conditioned on noise scheduling coefficient ${\overline{\alpha}}_t$, which increases with denoising steps, causing $r(t)$ to progressively decrease. $k$ controls the decrease rate, and $t_{thres}$ determines the inflection point of decreasing curve. $\text{Bernoulli}(r)$ represents the Bernoulli mask with probability $r(t)$. 
Empirically, applying feature removal to both query and key features in self-attention yields optimal performance (detailed analysis in Supp.\textcolor{blue}{6}): 
\begin{equation}
\begin{aligned}
\mathcal{A}^i &= \text{Softmax}(\frac{\text{Re}(\mathcal{Q}^i,\mathcal{M}^i_{con}) \cdot \text{Re}(\mathcal{K}^i,\mathcal{M}^i_{con}) }{\sqrt d}^T). \\
\end{aligned}
\end{equation}

\begin{figure}[t]
\centering
\includegraphics[width=1\columnwidth]{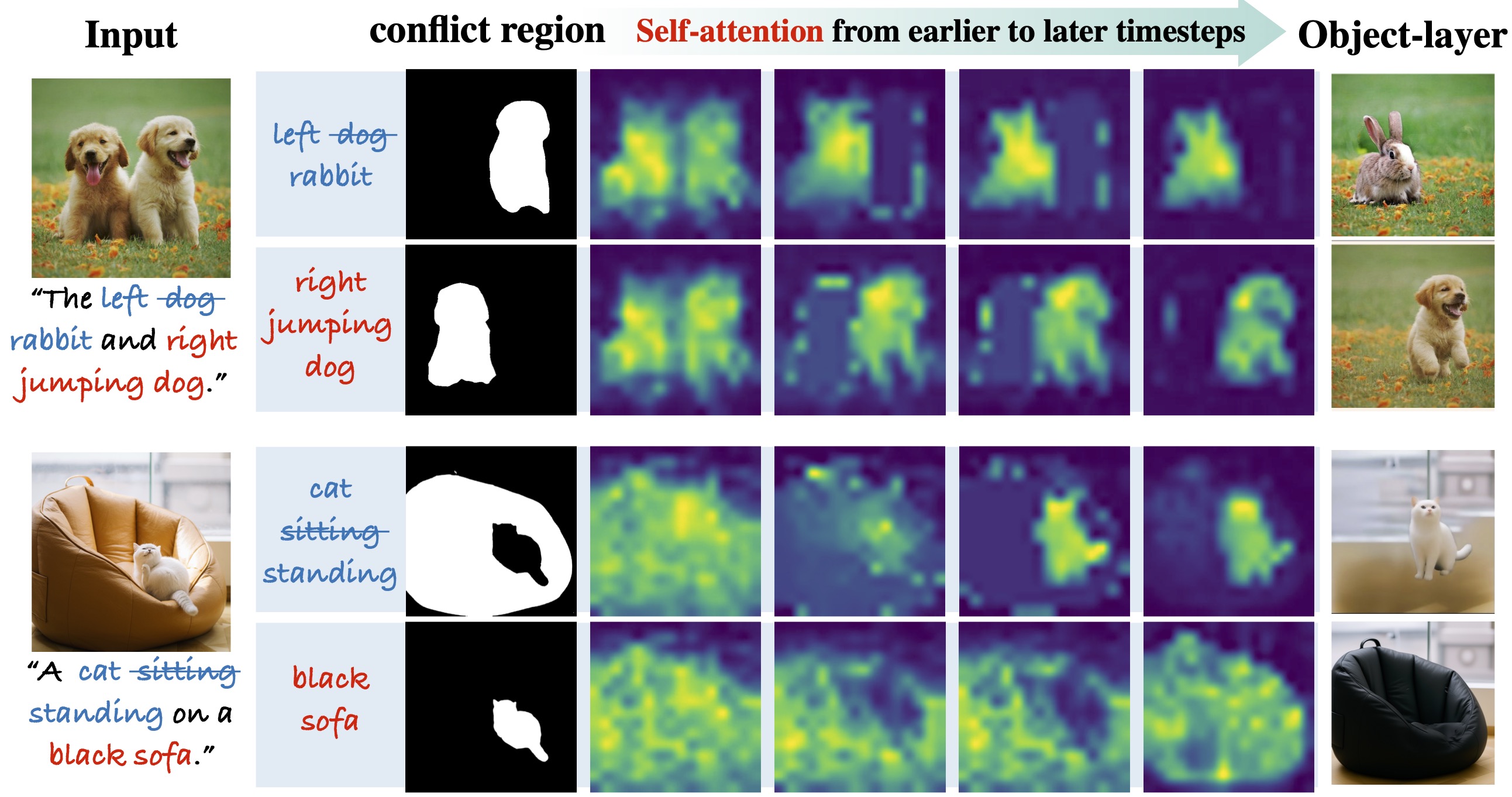} 
\caption{
 {Visualization} of attention features from earlier to later timesteps, driven by time-dependent region removing.
}
\label{fig-trr}
\end{figure}

The design rationale behind it is: modulating the feature removal intensity of conflict region conditioned on time-dependent image information density (weighted by $ \text{SNR}$). 
The visualization in Fig.\textcolor{blue}{\ref{fig-trr}} validates the effectiveness of this design: (1) in early timesteps, with severe conflict interference, $r(t)$ tends to $1$ to perform near-complete removal on conflict regions, preventing their leakage into subsequent editing.
(2) In later timesteps, as conflict object features are suppressed,  $r(t)$ tends to $0$ for minimal process on conflict region, ensuring adaptive and harmonious complementation of the surrounding background (\textit{e.g.},  ``dog's grassland") or obscured objects (\textit{e.g.}, ``sofa"), thereby supporting precise and scalable object editing on an exclusive layer.

\subsection{Object-layered Editing}
\label{editing}
To fully leverage the potential of multi-layer architecture, 
we further design intra-layer text guidance and cross-layer feature mapping, enabling compatible improvements for both textual semantic editing and geometric structural editing.

\noindent
\textbf{Textual Semantic Editing. } 
We extend the diffusion network into an $N+2$ multi-diffusion network ${\mathcal{L}=\{L_0,..., L_{N+1}\}}$, by reformulating the feature updating process (Eq.\textcolor{blue}{{\ref{Eq-gen}}}) with object-layered text guidance:
\begin{align}
 \hat{\phi}^i({z}^i_t, \mathcal{T}_e^i) = \mathbf{\mathcal{{A}}}^i\mathbf{\mathcal{{V}}}^i , \  \  for \  i \in [0, N+1], \\
  \mathcal{T}_e^i =\mathcal{T}_e - \bigcup{}_{\{ j|\mathcal{M}_{o}^j  \ in \  \mathcal{M}_{con}^i \}} ({O}_e^i ), \ for \  i \in [1, N]. 
\label{Eq9}
\end{align}
The layer $L_0$ indicates the source image reconstruction layer with corresponding conflict mask $\mathcal{M}_{con}^0=\emptyset$ and $\mathcal{T}_e^0= \mathcal{T}_s$. 
The layer $L_{N+1}$ is the canvas layer with conflict mask $\mathcal{M}_{con}^{N+1}= \bigcup_{i=1}^N(\mathcal{M}_{o}^i)$ and  $\mathcal{T}_e^{N+1}= \emptyset$ to generate image background for fusion.
Intuitively, the text guidance of each object layer is obtained by removing the object tokens located within its conflict region $\mathcal{M}_{con}^i$ 
from the global editing prompt $\mathcal{T}_e$, ensuring natural disentangled object editing guided by text. 
To simplify the subsequent discussions, we abbreviate the per-layer' spatial features $\hat{\phi}^i({z}^i_t, \mathcal{T}_e^i)$ as $\hat{\phi}^i$.

  \begin{figure}[!t]
\centering
\includegraphics[width=1\linewidth]{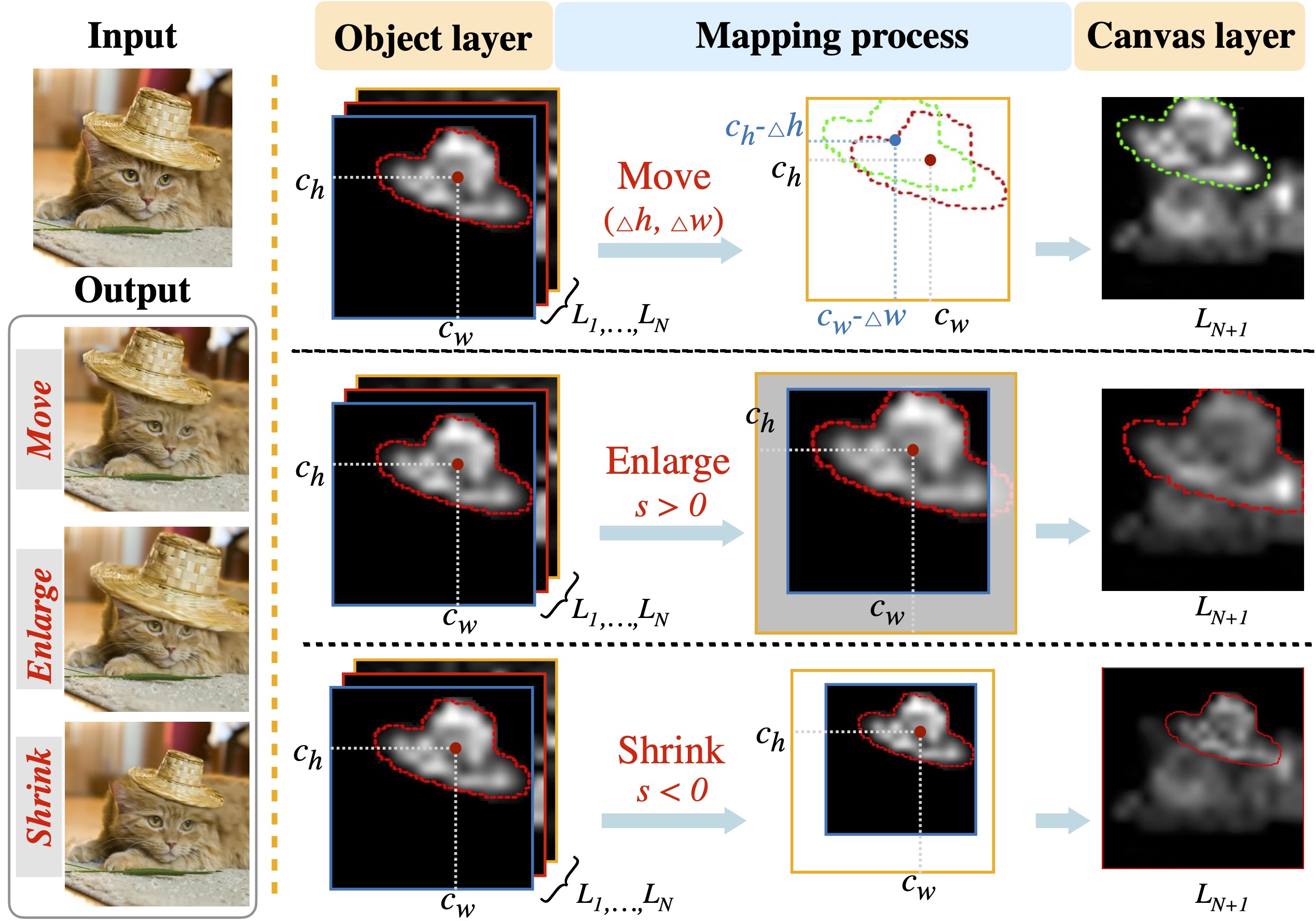} 
\caption{
{Diagram of \textit{centroid-aligned mapping principle}}. ($c_h,c_w$) is object centroid. Displacement $(\triangle h, \triangle w)$ and scale $s$ are additional input controls for moving/resizing.
}
\label{ge-method}
\end{figure}

\noindent
\textbf{Geometric Structural Editing. } 
\textit{LayerEdit} further supports geometric transformation, with a key insight: the multi-layer diffusion architecture inherently facilitates \textit{multi-step}, \textit{fine-grained} geometric feature mapping from reference object layers ($L_i$) to output canvas layer ($L_{N+1}$).
Conventional diffusion frameworks often suffer from source feature overwriting during mapping, limiting them to coarse-grained single-step operation \cite{schouten2025poem} or multi-step cumulative errors \cite{epstein2024diffusion}.
\textit{LayerEdit} overcomes these by maintaining persistent reference-to-canvas correspondence across all denoising timesteps, enabling iterative refinement of geometric transformations via multi-step attention-based feature mapping. 
Our framework explores two critical geometric editing functions: object moving and resizing, following the centroid-aligned mapping principle applied to the attention output features $\hat{\phi}^i$.
Fig.\textcolor{blue}{{\ref{ge-method}}} illustrates the diagram of centroid-aligned mapping principle within our multi-layer architecture via a representative multi-object case. More detailed formulas are provided in Supp.\textcolor{blue}{2}.

All textual and geometric editing integrates seamlessly, as intra-layer textual guidance and cross-layer geometric mapping operate within disentangled operation domains.

\subsection{Transparency-guided Layer Fusion}
\label{fusion}
Object-layered editing enables individual objects to extend beyond their original boundaries into adjacent objects or background regions, creating potential region overlapping. 
To ensure coherent overlapping layer fusion, we draw inspiration from image matting techniques \cite{yao2024vitmatte} and introduce \textit{transparency} learning to capture more precise inter-object structure guidance.
The transparency $\{\tau^i\}_{i=1}^N$ represents the contribution weights of object layers within each image patch. Object-layered fusion is performed on attention features in the canvas layer for each timestep: 
\begin{align}
 \hat{\phi}^{N+1}_{fusion}= \sum_{i=1}^{N}  \tau^i \odot \hat{\phi}^{i} + (1 -\sum_{i=1}^{N}  \tau^i ) \odot \hat{\phi}^{N+1}.
 \end{align}
 To achieve effective layer fusion, we formulate a novel constraint loss to calculate the optimal approximation of $\tau$:
\begin{gather}
 \mathcal{L}(\tau) = (( \hat{\phi}^{N+1}_{fusion}-   \hat{\phi}^{0}) \odot \mathcal{M}_\tau)^2+ \sum_{i=1}^{N} (\text{max}(0,-\tau^i ))^2   \nonumber \\
 +((1-\sum_{i=1}^{N}\tau^i)\odot \mathcal{M}_\tau)^2,
\end{gather}
where $\mathcal{M}_\tau$ marks the overlapping region, \textit{i.e.}, the $\tau$ of two layers are both greater than $0$.
Intuitively, the first term ensures the overlapping regions of fusion global image maintaining the spatial characteristics of source image,  thereby guiding transparency to capture inter-object structure relationship.
Physically,
the second term imposes a non negativity constraint on $\tau$, 
and the third term further constrains the global sum of overlapping layer region should be $1$. 
We solve this minimization problem via gradient descent:
\begin{gather}
\frac{d}{d\tau^n} \mathcal{L}(\tau) = 2 ( \hat{\phi}^{N+1}_{fusion}  -   \hat{\phi}^{0}) \odot \mathcal{M}_\tau \odot (\hat{\phi}^{n} - \hat{\phi}^{N+1} ) \nonumber \\
\ - 2  \text{max}(0,-\tau^n ) -2(1-\sum_{i=1}^{N}\tau^i)  \odot \mathcal{M}_\tau . 
\end{gather}
$\tau^i$ is initialized by object mask $\mathcal{M}^i_{o}$.
As $\tau$ is independent transparency weights of denoising network, it quickly converges within iterative inference timesteps.
This transparency learning, conditioned on source image spatial features, effectively captures inter-object structural relationships, promoting harmonious and coherent layer fusion.

\begin{table}
\fontsize{9}{11}\selectfont 
\setlength{\tabcolsep}{0.7mm}
  \centering
   {
 \begin{tabular}{lccccc}
    \toprule
   \multicolumn{1}{l}{\multirow{2}{*}{\textbf{Method}}}   & \multicolumn{1}{c}{\textbf{\textit{editability}}}                           &\multicolumn{2}{c}{\textbf{\textit{fidelity}}}           & \multicolumn{2}{c}{\textbf{\textit{quality}}}        \\ 
  &\textbf{CLIP-T}$^\uparrow$ &\textbf{CLIP-I}$^\uparrow$&\textbf{LPIPS}$^\downarrow$  &\textbf{FID}$^\downarrow$   &\textbf{KID}$^\downarrow$  \\
    \midrule
        InfEdit$_{,LCM}$ & 0.2253&0.8105&0.2529&80.62& 61.36 \\
       TurboEdit$_{,SDXL}$ &0.2147&0.8128&0.2491&90.16& 73.27 \\
      InstructCLIP$_{,IP2P}$ &0.2314&0.8214 &0.2307&76.27&62.40 \\
       \hdashline
   OIR$_{,SDv1.4}$&0.2538&0.8205& 0.2051&78.92&62.39\\
   GenArtist$_{,SDXL}$&0.2372&0.8028&0.2205&88.64&66.17\\
    ParallelEdits$_{,LCM}$ &0.2457&0.8108 &0.1924&74.47&57.81  \\
     LoMOE$_{,SDv2.0}$& 0.2506 &0.8365&0.1592&81.70&60.35 \\
    h-Edit$_{,SDv1.4}$& 0.2478 & 0.8216&0.1815&68.38& 52.28 \\
         \hdashline
\textbf{Ours}$_{,SDXL}$&\underline{0.2762}&\textbf{0.8420}&\textbf{0.1358}&\underline{60.13}&\underline{45.29} \\
    \textbf{Ours}$_{,FLUX}$&\textbf{0.2857}&\underline{0.8408}&\underline{0.1372}&\textbf{57.91}&\textbf{39.72} \\
    \bottomrule
  \end{tabular}}
  \caption{{Quantitative comparisons} with existing methods. 
  }
  \label{all}
\end{table}

\section{Experiments}

\subsection{Experimental Setups}

\begin{figure*}[!t]
\centering
\includegraphics[width=0.97\linewidth]{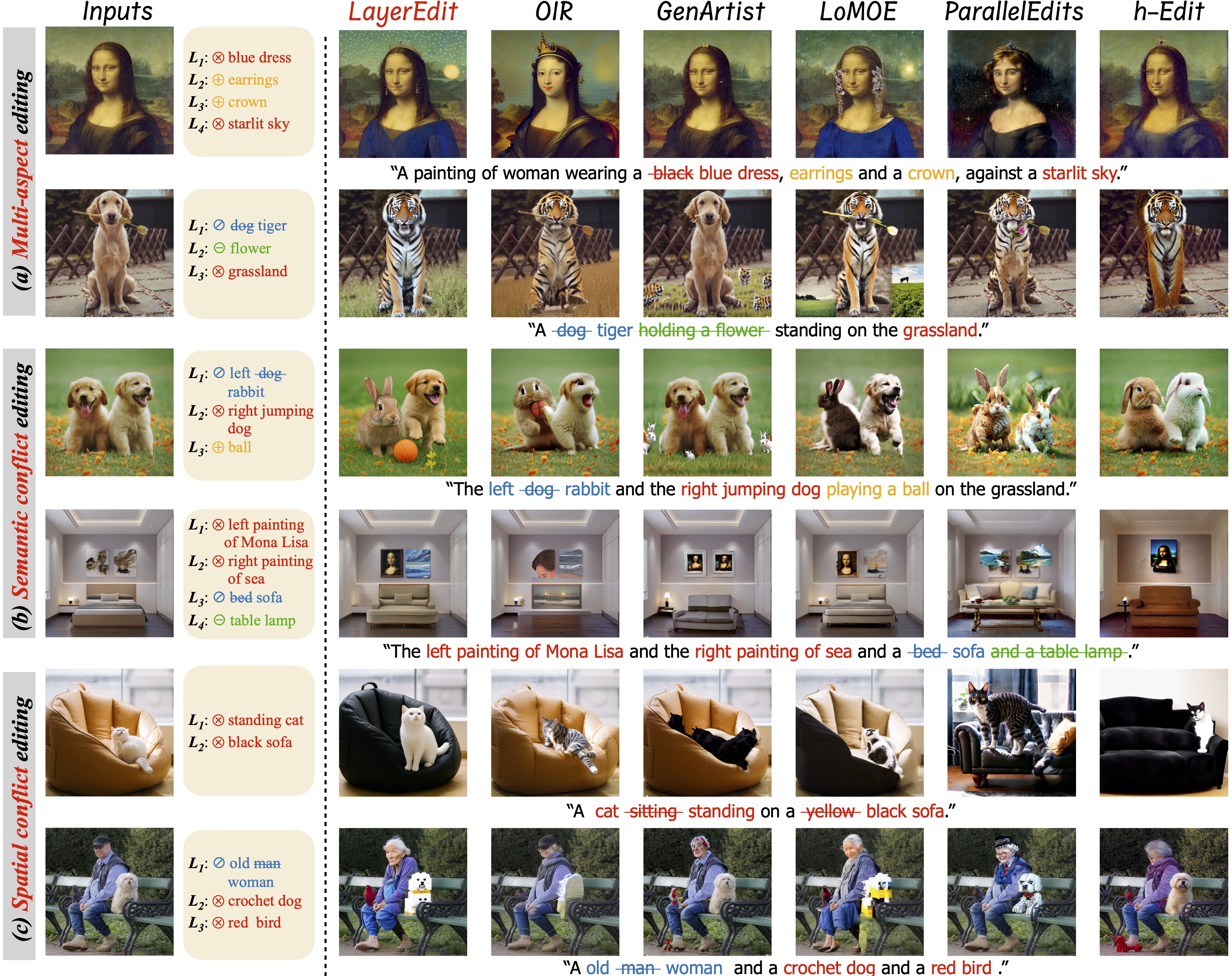} 
\caption{{Qualitative comparison} with SOTAs. $\{${$\oslash$}, {$\oplus$},  {$\ominus$}, {$\otimes$}$\}$ denote different editing operations defined as Sec.\textcolor{blue}{\ref{Qualitative}}.
\textit{LayerEdit} exhibits unprecedented editability for multi-aspect/object editing. 
More comparisons with industrial-grade models in Supp.\textcolor{blue}{4}. 
} 
\label{fig-SOTA}
\end{figure*}


\noindent
\textbf{Implementation and Datasets.}
As a plug-and-play framework, \textit{LayerEdit} can be easily integrated into existing backbones.
We adopt Stable Diffusion XL \cite{podell2023sdxl} as main backbone for a fair comparison, and FLUX \cite{flux2024} to further validate generalization (detailed in Supp.\textcolor{blue}{3}).
We use DDIM sampler with $50$ sampling steps and classifier-free guidance scale of $7.5$.
Experimentally, we obtain one of optimal settings with IoU threshold $\eta=0.3$, rate $k=5$, timestep threshold $t_{thres}$ at $20$ (query) and $40$ (key).
All experiments are conducted on two typical multi-object datasets: OIR-bench \cite{yang2024object} and LoMOE-bench \cite{chakrabarty2024lomoe}. 
Notably, the proposed multi-layer diffusion framework with parallelized layer-wise computation, incurs almost \textbf{\textit{no additional time costs}}, enabling efficient editing of more than $6$-object on a single A40 GPU.

\noindent
\textbf{Baselines.}
We compare with state-of-the-art (SOTA) baselines for TMOE: h-Edit \cite{nguyen2025h}, ParallelEdits \cite{huang_etal_neurips24c}, LoMOE \cite{chakrabarty2024lomoe},  GenArtist \cite{wang2024genartist}, and OIR \cite{yang2024object}. For comprehensive evaluation,
other SOTA general editing methods (InfEdit, \cite{xu2024inversion}; TurboEdit, \cite{wu2024turboedit}; InstructCLIP, \cite{chen2025instruct}) are also compared, processed as iteratively multi-turn editing for multi-object. 
%

\noindent
\textbf{Evaluation Metrics.}
\textbf{\textit{Image fidelity}}: 
we evaluate with both CLIP embedding similarity  (CLIP-I, \cite{radford2021learning}) and learned perceptual image patch similarity (LPIPS, \cite{zhang2018unreasonable}).  \textbf{\textit{Editability}}: we evaluate with CLIP embedding similarity between edited prompts and images (CLIP-T). \textbf{\textit{Image quality}}: we evaluate with Frechet Inception Distance (FID, \cite{heusel2017gans}) and Kernel Inception Distance (KID, \cite{lucic2018gans}).

\subsection{Main Results}

\noindent
\textbf{Quantitative Results. }
As illustrated in Tab.\textcolor{blue}{\ref{all}}, \textit{LayerEdit} outperforms existing methods in all metrics, improving editability ($\mathbf{12.6\%}$ in CLIP-T), fidelity ($\mathbf{14.7\%}$ in LPIPS) and global quality ($\mathbf{15.3\%}$ in FID).
These significant improvements validate the distinctive superiority of our multi-layer disentangled editing paradigm, enabling high-quality intra-object modification with seamless inter-object coherence. 

\noindent
\textbf{Qualitative Results.}
\label{Qualitative}
Fig.\textcolor{blue}{\ref{fig-SOTA}} showcases \textit{LayerEdit}'s revolutionary capabilities:
(1) \textbf{\textit{Wide spectrum of editing}}, supporting
 object \textit{replacement} ({$\oslash$}), \textit{addition} ({$\oplus$}), \textit{removal} ({$\ominus$}) and \textit{appearance/posture/background manipulations} ({$\otimes$}, treating background as a regular object layer). 
Furthermore, \textit{LayerEdit} also effectively supports the \textit{multi-aspect editing}, through treating objects' aspects (\textit{e.g.}, Fig.\textcolor{blue}{\ref{fig-SOTA}.a}, woman's ``dress" and ``earrings" ) as independent objects.
(2) \textbf{\textit{Inter-object conflict resolution}} across complex multi-object scenarios, enabling
 precise text-guided editing for semantically related objects (\textit{e.g.}, Fig.\textcolor{blue}{\ref{fig-SOTA}.b}, multi-directional ``dog" or ``painting") and spatially overlapping objects (\textit{e.g.}, Fig.\textcolor{blue}{\ref{fig-SOTA}.c}, occluded ``cat" or ``sofa"), with seamless and harmonious global texture. 
This conclusion verifies the effectiveness of multi-layer disentangled framework for unconstrained intra-object editing and coherent inter-object fusion,
while existing baselines suffer from either 
strict intra-object boundaries (\textit{e.g.}, LoMOE's artifacts) or inter-object attention leakage (\textit{e.g.}, ParallelEdits' mis-editing on two ``dog/painting").

\subsection{Results on Geometric Editing} 
\begin{figure}[!t]
\centering
\includegraphics[width=0.98\linewidth]{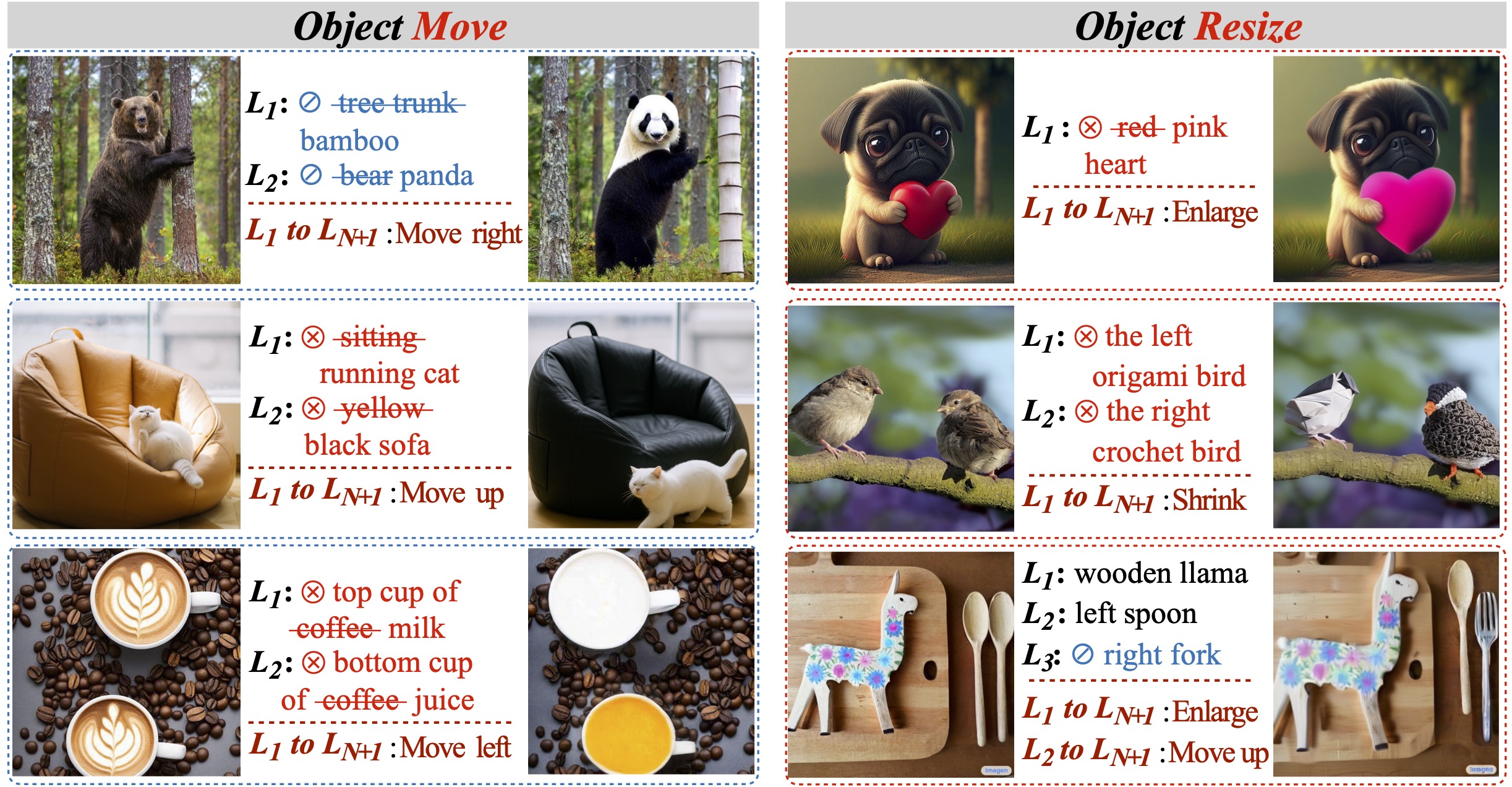} 
\caption{
{Qualitative results} of geometric editing. 
}
\label{fig-geometry-all}
\end{figure}

As shown in Fig.\textcolor{blue}{\ref{fig-geometry-all}}, \textit{LayerEdit} demonstrates excellent geometric editing superiority, including:
(1) \textbf{\textit{Seamless dual-application}} of text- and geometric-guided modifications to single or multiple objects without constraints or conflicts, which originates from our multi-layer framework effectively disentangles and coordinates the operational domains of both editing functions.
(2) \textbf{\textit{Adaptive occlusion completion}} (\textit{e.g.}, $1$-st row, occluded ``heart" or ``bamboo"), 
 demonstrating \textit{LayerEdit}'s unique capability in region-unconstrained and scale-adaptive object manipulation.
More qualitative results with SOTA geometric editing methods refer to Supp.\textcolor{blue}{5}. 

\subsection{Visualization Results of Object Layers}

\begin{figure}[!t]
\centering
\includegraphics[width=0.97\linewidth]{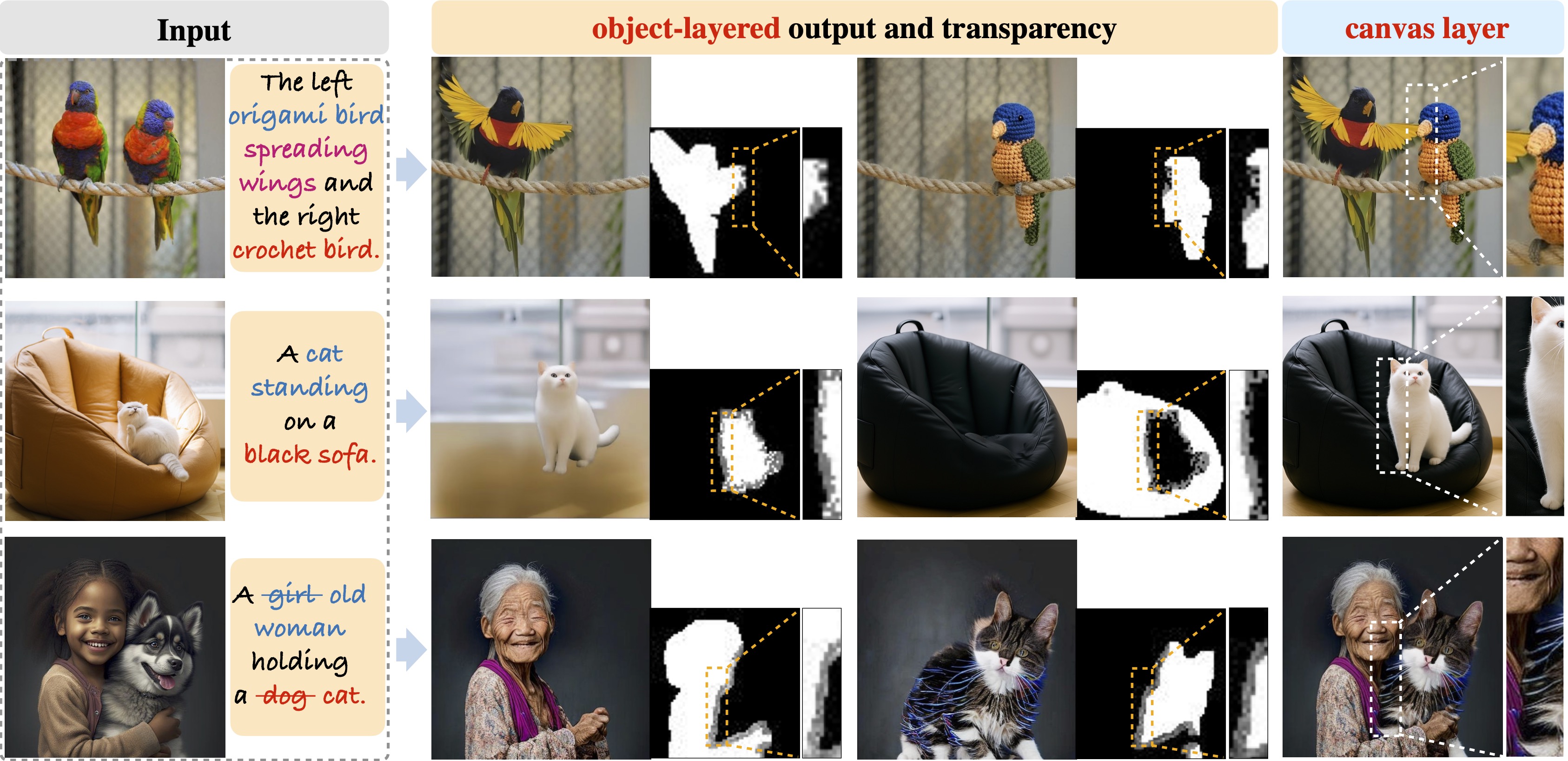} 
\caption{
{Visualization results} of object-layered output and transparency. Boxes \textit{highlight} the overlapping regions.
}
\label{fig-ablation}
\end{figure}

Fig.\textcolor{blue}{\ref{fig-ablation}} visualizes the object-layered outputs and transparency maps.
Intuitively, (1) for \textbf{\textit{layer decomposition}}, \textit{LayerEdit} achieves precise removal of conflict regions (\textit{e.g.}, semantically/spatially conflicting ``bird"/``sofa") 
with adaptive region completion,
thereby enabling unconstrained \textbf{\textit{object-layered editing}}
 with arbitrary structural modifications and expansions (\textit{e.g.}, ``bird spreading wings" and ``cat standing") within exclusive layer.
(2) For \textbf{\textit{layer fusion}}, spatial features conditioned transparency learning effectively captures inter-object structure (\textit{e.g.}, ``cat on sofa", ``woman hold cat"), further providing coherent overlapping region fusion for region-scalable object editing.
These results verify that the 
\textbf{\textit{essence of LayerEdit}} is to fundamentally model a precise decompose-editing-fusion pipeline, to achieve truly disentangled multi-object editing, free from both intra-object region constraints and inter-object conflicts.

\begin{table}
  \centering
  \fontsize{9}{11}\selectfont 
\setlength{\tabcolsep}{0.7mm}
  {
  \begin{tabular}{ccccc|cc}
    \toprule
    \multicolumn{1}{c}{\multirow{1}{*}{\textbf{ID}}}   &  \multicolumn{1}{c}{\multirow{1}{*}{\textbf{Settings}}}                &    \multicolumn{1}{c}{\multirow{1}{*}{\textbf{ CLIP-T }$^\uparrow$ }}      &   \multicolumn{1}{c}{\multirow{1}{*}{\textbf{ CLIP-I }$^\uparrow$ }}          \\ 
    \midrule
   \textbf{D-v1} &w/o $t$-dependent $r$, w/ fixed $r$ &0.2386&0.8103\\
      \textbf{D-v2} &w/o $t$-dependent $r$, w/ linear $r$ &0.2560&0.8194\\
\hdashline
  \textbf{E-v1} &w/o intra-layer textual guidance & 0.2489 &0.8045 \\
\hdashline
   \textbf{F-v1} &w/o  transparency, w/ mask  &0.2415&0.8352\\
      \textbf{F-v2} &w/o 1-st structure loss term &0.2536 &0.8216 \\
            \textbf{F-v3} &w/o 2-nd non-negativity loss term & 0.2508 &0.8171      \\
    \textbf{F-v4} &w/o 3-th normalization loss term & 0.2692 &0.8375      \\
    \midrule
         \textbf{-- --} & full model & \textbf{0.2762} &\textbf{0.8420} \\ 
  \bottomrule
\end{tabular}}
  \caption{
   Ablation on effectiveness of different components.
  } 
    \label{ablation}
\end{table}

\subsection{Ablation Studies}
We further ablate on model components to verify that our \textit{LayerEdit} achieves one of the optimal designs for this decompose-editing-fusion architecture. 
We compare with: 
1) D-v1/v2: model without \textbf{t}ime-dependent {r}egion {r}emoving but uses fixed (optimal at $r=0.8$) or linear decreasing removal intensity ($r$ from $1$ to $0$) on conflict region; 
2) E-v1: model without intra-layer textual guidance but all use $\mathcal{T}_e $ in Eq.\textcolor{blue}{\ref{Eq9}}.
3) F-v1:  model without transparency learning but with mask-guided layer fusion;
4) F-v2/3/4: model without the $1/2/3$-th loss term in transparency learning.

As reported in Tab.\textcolor{blue}{\ref{ablation}}, we conclude that: 
(1) for decomposition, time-dependent region removing strategy excels at conflict region constraint, leading to conflict-free object-layered decomposition and editing, with a remarkable $15.8\%$ improvement in CLIP-T.
(2) For editing, the tailored intra-layer textual guidance better supports the multi-layer architecture for natural disentangled editing.
(3) For fusion, 
model without transparency guidance severely impairs editability (CLIP-T dropped by $14.4\%$), proving transparency learning is essential for structure-coherent fusion in region-scalable object editing. 
 More ablation results in qualitative comparisons, and aware/removal parameters $\eta, k, t_{thres}$ in Supp.\textcolor{blue}{6}.


\section{Conclusion}
In this paper, we present \textit{LayerEdit}, a training-free multi-layer disentanglement editing framework which, for the first time, through precise object-layered decomposition and fusion, enhances text-driven multi-object editing free from inter-object conflicts and intra-object constraints.
Different from existing methods fixated on intra-object localize-editing, our method shifts the focus into inter-object interactions, by explicitly identifying and constraining inter-object conflict regions to achieve \textit{conflict-aware layer decomposition}, and through inter-object structural modeling to facilitate structure-coherent \textit{transparency-guided layer fusion}. 
Moreover, the tailored \textit{object-layered editing} module further supports the constructed multi-layer architecture, establishing novel intra-layer textual guidance and cross-layer feature mapping, to achieve comprehensive improvements for arbitrary semantic and structural editing functions.
Extensive experiments demonstrate the superiority of our \textit{LayerEdit}.

\bibliographystyle{abbrv}
\bibliography{aaai26}

\clearpage

\appendix


\begin{figure*}[!t]
\centering
\includegraphics[width=1\linewidth]{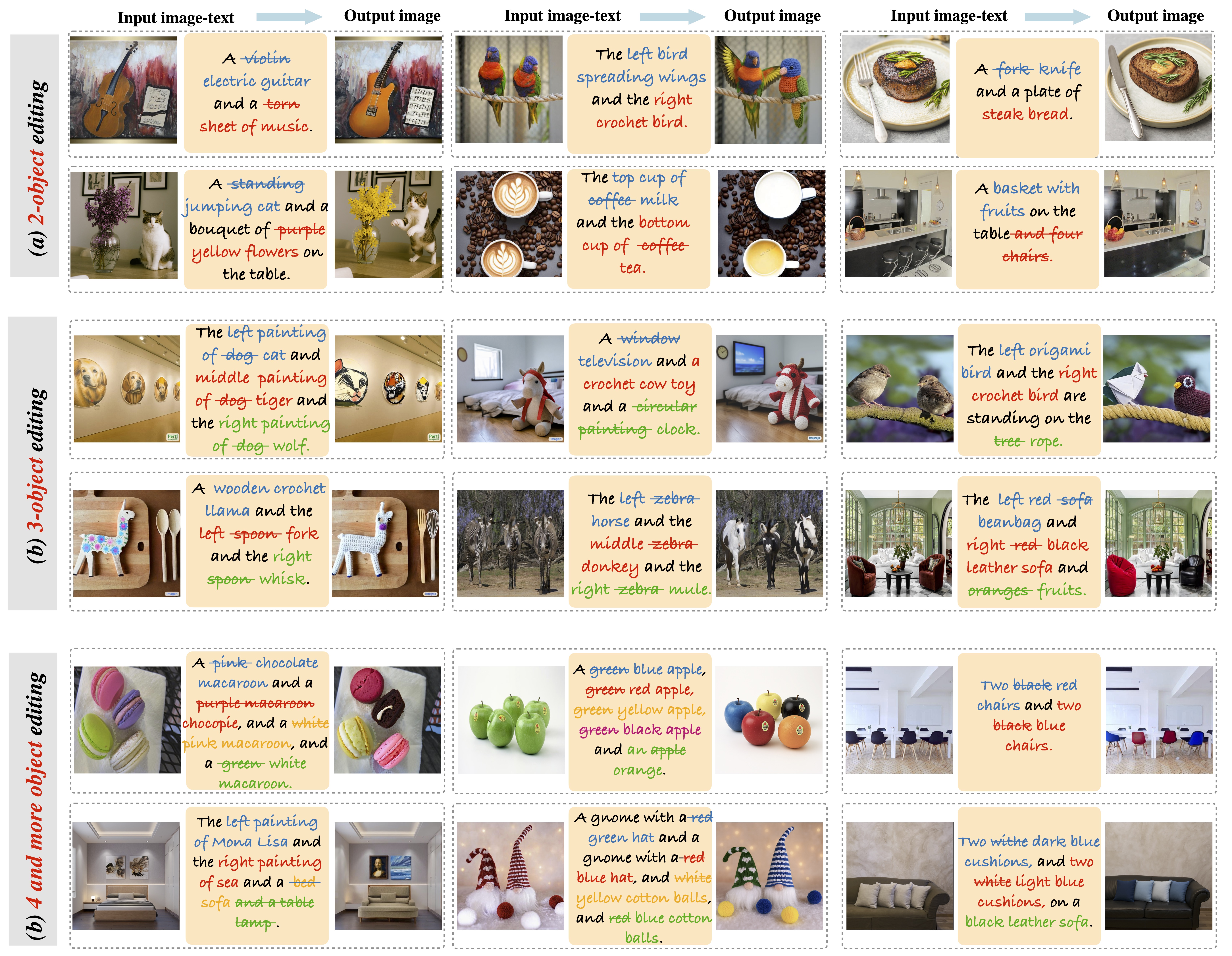} 
\caption{{More Qualitative results} on multi-object editing, show that \textit{LayerEdit} can generate high-quality edited images in  complex multi-object scenarios.
} 
\label{MOE-more}
\end{figure*}

\section{More Results of Multi-Object Editing}

More visualization results of editing $2$ objects, $3$ objects, and more than $4$ objects are provided in Fig.\textcolor{blue}{\ref{MOE-more}}, highlighting the powerful editing performance of \textit{LayerEdit} in complex multi-object scenarios.

\section{Details of Geometric Structural Editing}
\label{ge-formule}
In this section, we elaborate on the implementation of designed geometric structural editing, including object removing and resizing, which follows the  centroid-aligned mapping principle.

For object moving, given movement vectors $\langle \triangle h, \triangle w \rangle$ on $i$-th object, we employ the moving mapping at the attention output feature levels. Specifically:
\begin{align}
\text{Move}(x, \triangle h, \triangle w)  &=  x(h-\triangle h, w-\triangle w) \\\nonumber
\hat{\phi}^{N+1}( i,\triangle h, \triangle w)
=  &\hat{\phi}^{N+1}\odot  (1- \text{Move}({\tau}^i,\triangle h, \triangle w)) \\
 + \text{Move}(  \hat{\phi}^{N+1},\triangle h, &\triangle w) \odot \text{Move}({\tau}^i,\triangle h, \triangle w),
\end{align}
 where $\text{Move}(\cdot)$ indicates the moving operation. These operations successfully shift the $i$-th object centroid with displacement $\langle \triangle h, \triangle w \rangle$.
 
 For object resizing, we use spherical interpolation for feature resizing and map the resized features based on centroid alignment. Specifically, with a source image size $\langle H,W \rangle$, we define the object centroid and resizing operation as:
 \begin{align}
{c}^i &= \frac{1}{\sum_{h,w}\mathbf{\tau}^i_{h,w}}
    \begin{pmatrix}
\sum_{h,w} w \cdot \mathbf{\tau}^i_{h,w} \\
\sum_{h,w} h \cdot \mathbf{\tau}^i_{h,w} 
    \end{pmatrix} \\
    \text{Resize}(x, s)  &=  \text{Interpolation}(x,H \cdot s, W \cdot s) 
     \end{align}
Since the resized features and source features have different shapes, we construct the resizing mapping with centroid alignment as:
 \begin{align}
\hat{\phi}^{N+1}[{c}^i ,\hat{h},\hat{w}] =  \text{Resize}(\hat{\phi}^{i} \odot \tau^{i}, s)[{c}^i ,\hat{h},\hat{w}]  \\\nonumber
+ \hat{\phi}^{N+1}[{c}^i ,\hat{h},\hat{w}] \odot (1-\tau^{i})[{c}^i ,\hat{h},\hat{w}],\\
for \ \hat{h},\hat{w}=\text{min}(H,H\cdot s),  \text{min}(W,W\cdot s)
     \end{align}
where $x[c,h,w]$ indicates a matrix extracted from $x$ with centroid $c$, height $h$, and width $w$.
All geometric editing functions can be adapted to multi-object editing scenarios, combined with textual editing.

\section{More Details for Implementation}
\label{implementation}

The model is implemented in PyTorch with a single NVIDIA GeForce RTX A40 GPU. 
While adopting FLUX \cite{flux2024} as the backbone, we set the sampling steps to $30$ and the classifier-free guidance scale to $3.5$. The IoU threshold $\eta$ is $0.3$, and the rate $k$ is $5$, and timestep threshold $t_{thres}$ are $15$ (query) and $25$ (key).

\section{More Results with Industrial-grade Models}

\begin{figure}[!t]
\centering
\includegraphics[width=1\linewidth]{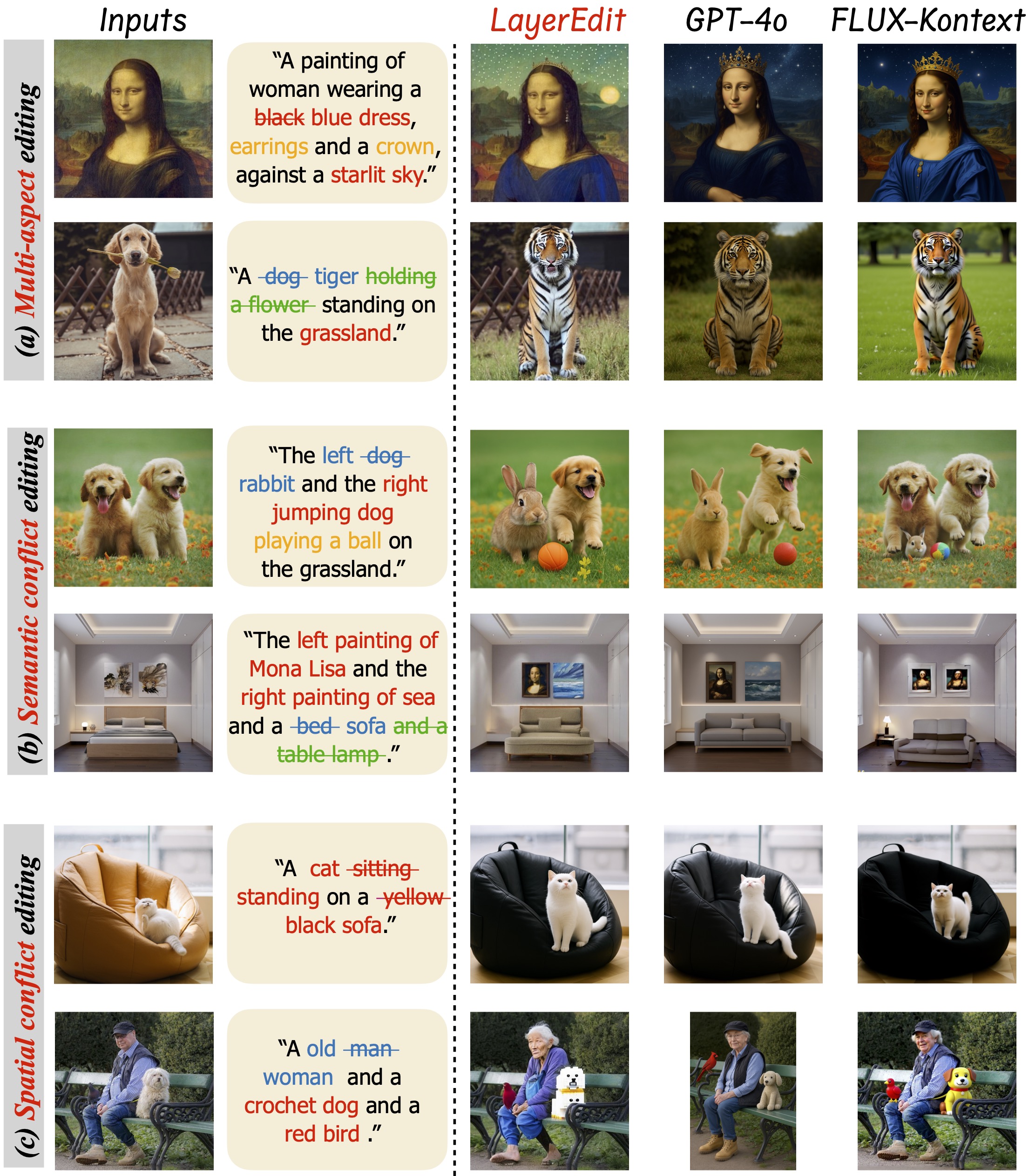} 
\caption{{Qualitative comparison} with SOTA industrial-grade models. 
} 
\label{industrial-SOTA}
\end{figure}

More comparison results with industrial-grade models GPT-4o \cite{2017automatic}, and FLUX Kontext \cite{labs2025flux1kontextflowmatching} are provided in Fig.\textcolor{blue}{\ref{industrial-SOTA}}.
Compared with these state-of-the-art methods in the field of image editing, our model demonstrates powerful editing capabilities that surpass FLUX Kontext (\textit{e.g.}, Fig.\textcolor{blue}{\ref{industrial-SOTA}}.b, FLUX Kontext fails to resolve semantic conflicts between multi-directional dogs and paintings) and outperforms GPT-4o in consistency preservation (\textit{e.g.}, Fig.\textcolor{blue}{\ref{industrial-SOTA}}.a, GPT-4o's over-editing on ``woman" texture and ``tiger" background). These results demonstrate that \textit{LayerEdit} has achieved powerful performance on par with industrial-grade models.

However, \textit{LayerEdit} still has limitations in terms of the resolution of generated images (\textit{e.g.},  the $4$-th row in Fig.\textcolor{blue}{\ref{industrial-SOTA}}, the generated ``Mona Lisa" is clearly less sharp than that generated by GPT-4o). This limitation stems directly from the used backbone rather than the multi-layer disengagement editing framework itself.



\section{More Results on Geometric Editing}

\begin{figure*}[!t]
\centering
\includegraphics[width=0.97\linewidth]{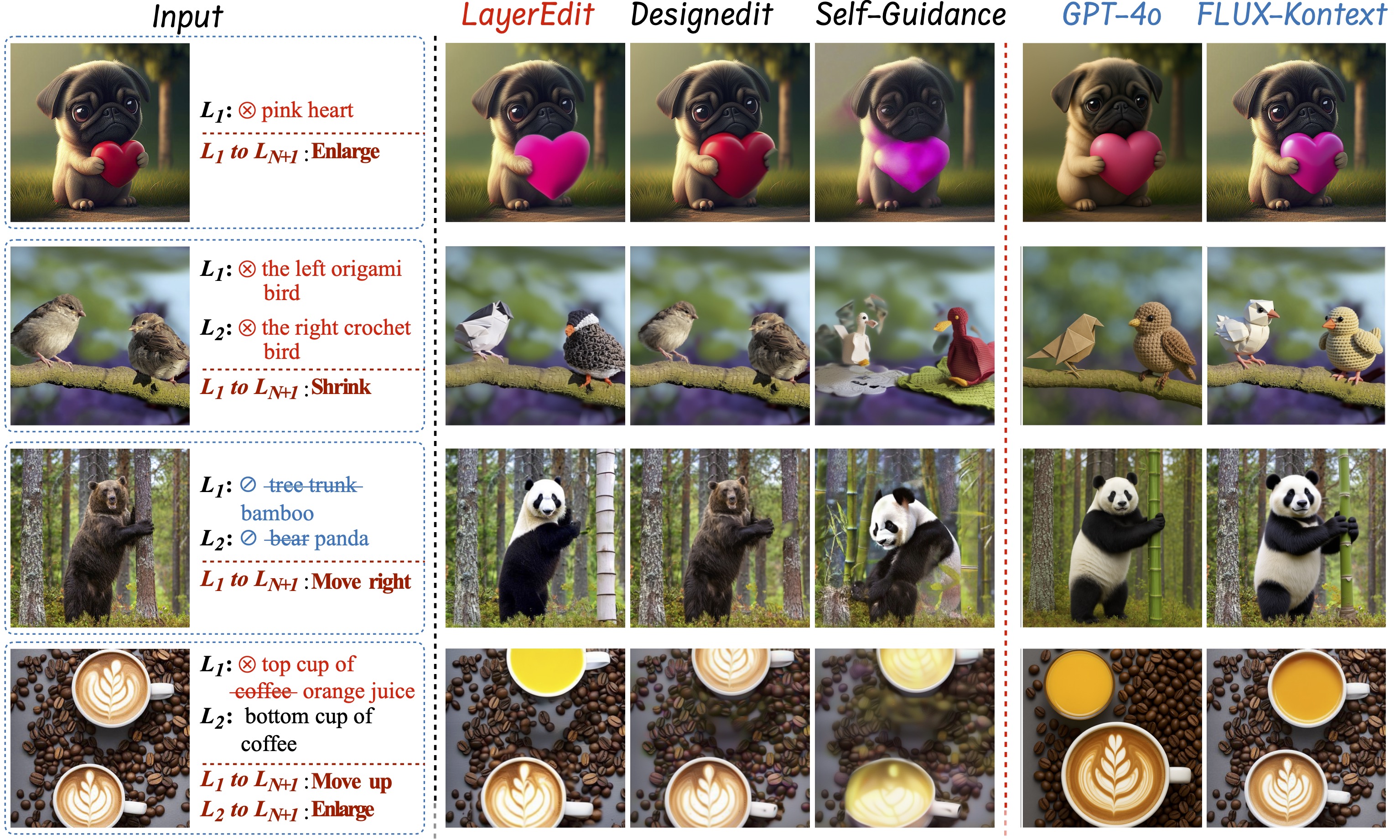} 
\caption{{Qualitative comparison} with SOTAs on geometric editing. 
} 
\label{ge-SOTA}
\end{figure*}

In this section, we present additional results on geometric editing and comparisons with state-of-the-art (SOTA) baselines. 
We compare \textit{LayerEdit} with leading works that are capable of geometric editing: DesignEdit \cite{jia2024designedit}, Self-Guidance \cite{epstein2024diffusion} and industrial-grade models GPT-4o \cite{2017automatic}, and FLUX Kontext \cite{labs2025flux1kontextflowmatching}. As shown in Fig.\textcolor{blue}{\ref{ge-SOTA}}, \textit{LayerEdit} demonstrates strong capabilities in two key aspects: (1) combined editing with both textual and geometric guidance; (2) adaptive and harmonious complementing of occluded regions. 
 These results further verify that \textit{LayerEdit} enables adaptive and scalable object editing without region constraints.

\section{More Ablation Studies}
\label{Ablation}

\begin{figure*}[!t]
\centering
\includegraphics[width=1\linewidth]{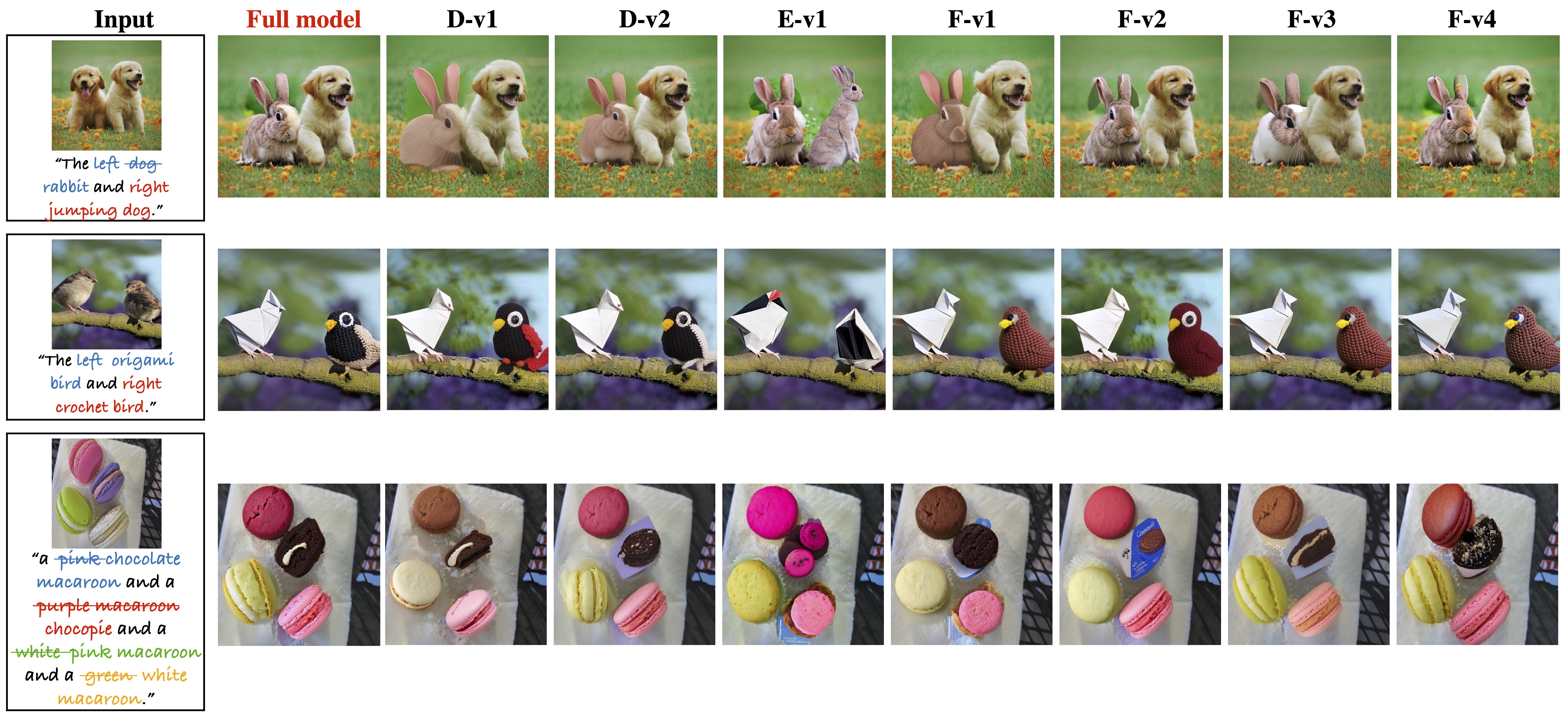} 
\caption{{Qualitative ablation studies} on different components.
} 
\label{ablation-components}
\end{figure*}

More qualitative results of the ablation studies are presented in Fig.\textcolor{blue}{\ref{ablation-components}},  further validating the effectiveness and rationality of the designed decompose-editing-fusion framework.
Furthermore, we conduct more detailed ablation studies on parameters $\eta, k, t_{thres}$, to further demonstrate the model's robustness, where the model can achieve high-quality editing across a relatively wide range of parameters without relying on fine parameter tuning.
All ablation experiments are conducted based on the principle of controlling variables.

\subsection{Ablation on Parameter $\eta$}
\begin{figure*}[!t]
\centering
\includegraphics[width=0.9\linewidth]{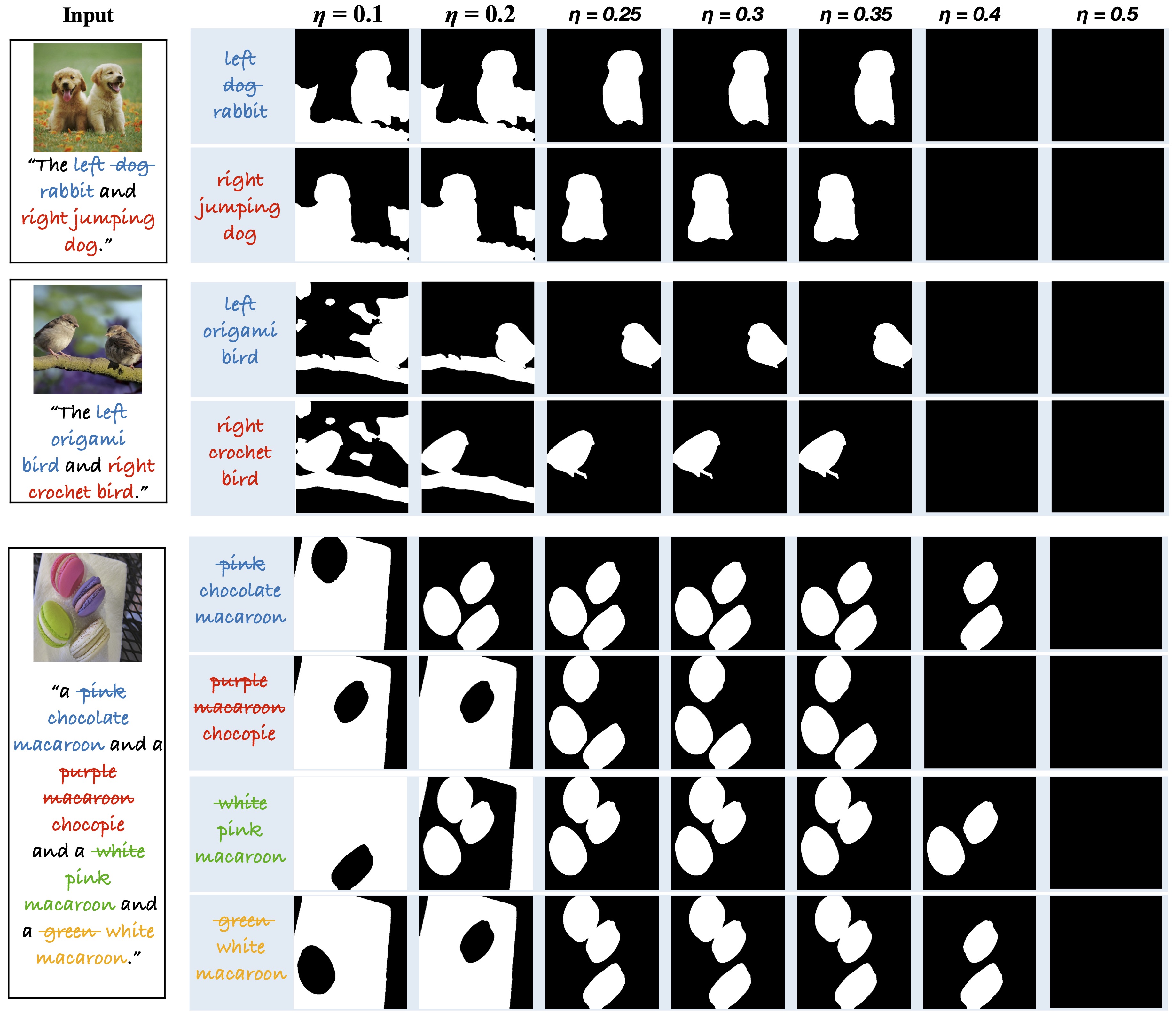} 
\caption{{Ablation results on parameter $\eta$}.
} 
\label{aware-ablation}
\end{figure*}

 The results are shown in Fig.\textcolor{blue}{\ref{aware-ablation}}. Within a proper range (red box marks, $\eta \in [0.25, 0.35]$), the model demonstrates the accurate identification of inter-object conflict regions, confirming the effectiveness and robustness of our conflict-aware multi-layer learning model.

 \subsection{Ablation on Parameter $k$}

 \begin{table}
  \centering
  \fontsize{9}{11}\selectfont 
\setlength{\tabcolsep}{1mm}
  {
  \begin{tabular}{ccccc|cc}
    \toprule
    \multicolumn{1}{c}{\multirow{1}{*}{\textbf{Settings}}}           &    \multicolumn{1}{c}{\multirow{1}{*}{\textbf{ CLIP-T }$^\uparrow$ }}      &   \multicolumn{1}{c}{\multirow{1}{*}{\textbf{ CLIP-I }$^\uparrow$ }}          \\ 
    \midrule
   $t_{thres}^Q = t_{thres}^K = 25, {k=1} $ &0.2386&0.8173\\
    $t_{thres}^Q = t_{thres}^K = 25,  {k=3} $&0.2562&0.8235\\
     $t_{thres}^Q = t_{thres}^K = 25, \mathbf{k=5} $ &\textbf{0.2585} & \textbf{0.8249}\\
 $t_{thres}^Q = t_{thres}^K = 25, {k=8} $ &0.2537&0.8223\\
     $t_{thres}^Q = t_{thres}^K = 25, {k=10} $&0.2483&0.8195\\

  \bottomrule
\end{tabular}}
  \caption{
   Ablation results on parameter $k$.
  } 
    \label{ablationK}
\end{table}

\begin{figure*}[!t]
\centering
\includegraphics[width=0.97\linewidth]{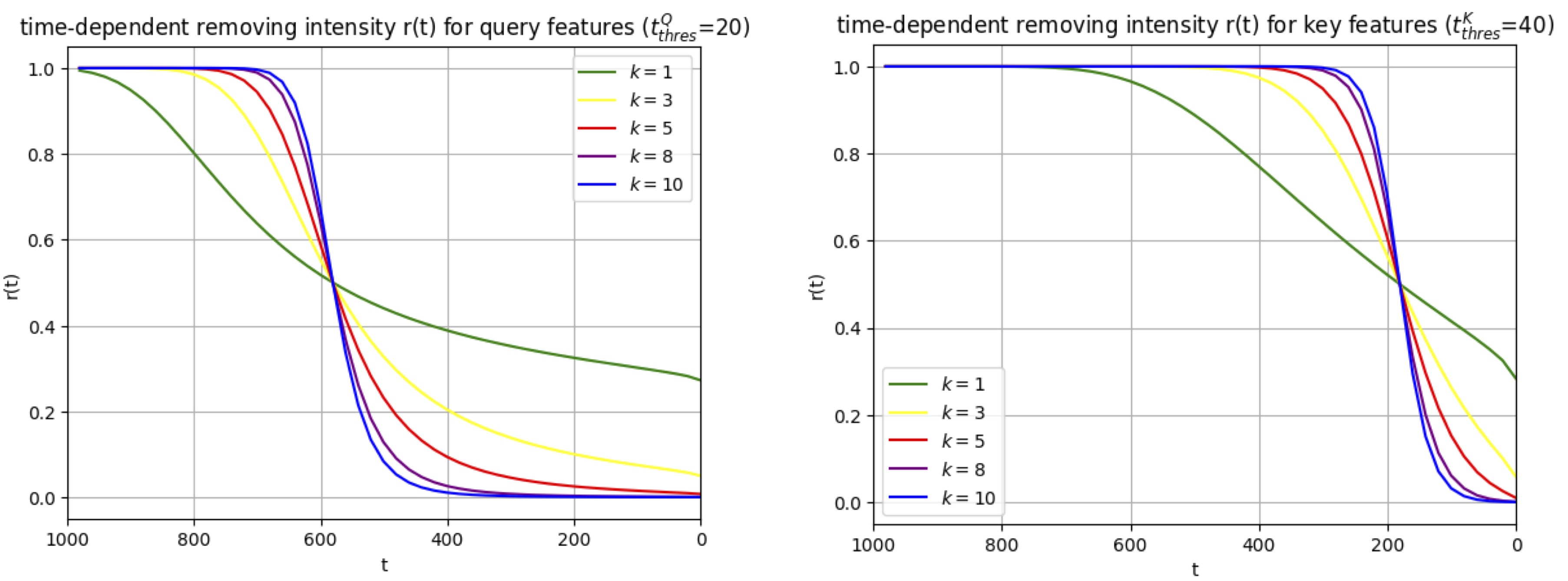} 
\caption{{Visualization results of $r(t)$ curves on different parameter $k$ settings.} 
} 
\label{kcurve}
\end{figure*}

 Considering that parameters $k$ and $t_{thres}$ jointly control the time-dependent region removing intensity $r(t)$, we explore a general setting for $k$ to reduce parameter complexity.
Specifically, We simply fixed $t_{thres}^Q = t_{thres}^K = 25$, which is half of the total number of denoising steps, and tested different parameter $k$ settings. The results in Tab.\textcolor{blue}{\ref{ablationK}} show that optimal performance is achieved at $k = 5$, with comparable performance maintained across a  proper range $k \in [3,8]$, also highlighting the model's robustness.
 
 We also visualized the $r(t)$ curves for different parameter $k$ settings, as shown in Fig.\textcolor{blue}{\ref{kcurve}}.
 This visualization more intuitively demonstrates that with the setting of $k=5$, the constructed SNR-weighted function $r(t)$ achieves a smooth transition from a higher intensity (approaching $1$) in early timesteps  to a lower intensity (approaching $0$) in late timesteps.



\subsection{Ablation on Parameter $t_{thres}$}

\begin{figure*}[!t]
\centering
\includegraphics[width=0.8\linewidth]{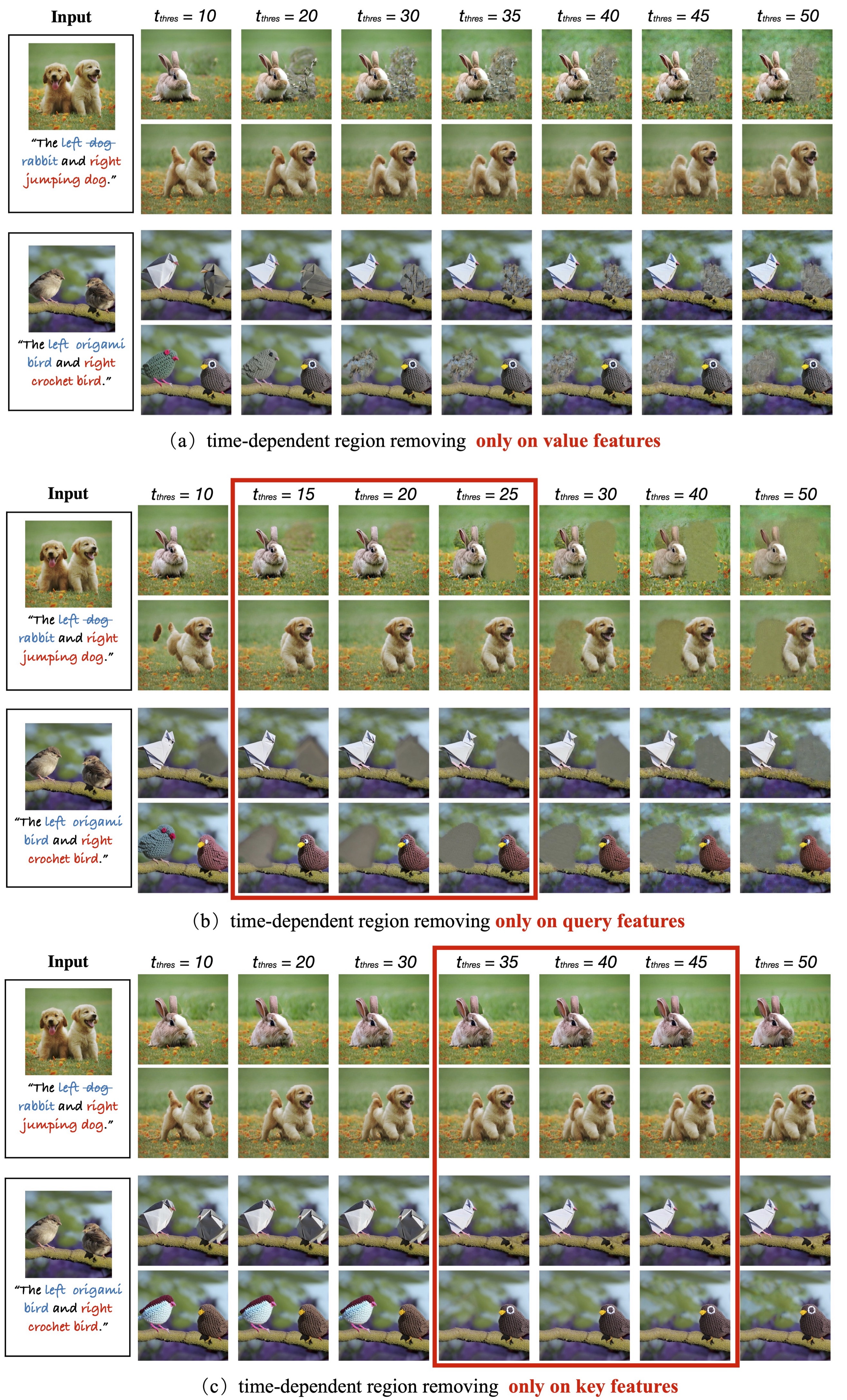} 
\caption{{Ablation results on parameter $t_{thres}$ with time-dependent region removing scheme acting only on self-attention's query $\mathcal{Q}$, key $\mathcal{K}$, and value $\mathcal{V}$ features.} 
} 
\label{QKV-ablation}
\end{figure*}

\begin{figure*}[!t]
\centering
\includegraphics[width=0.8\linewidth]{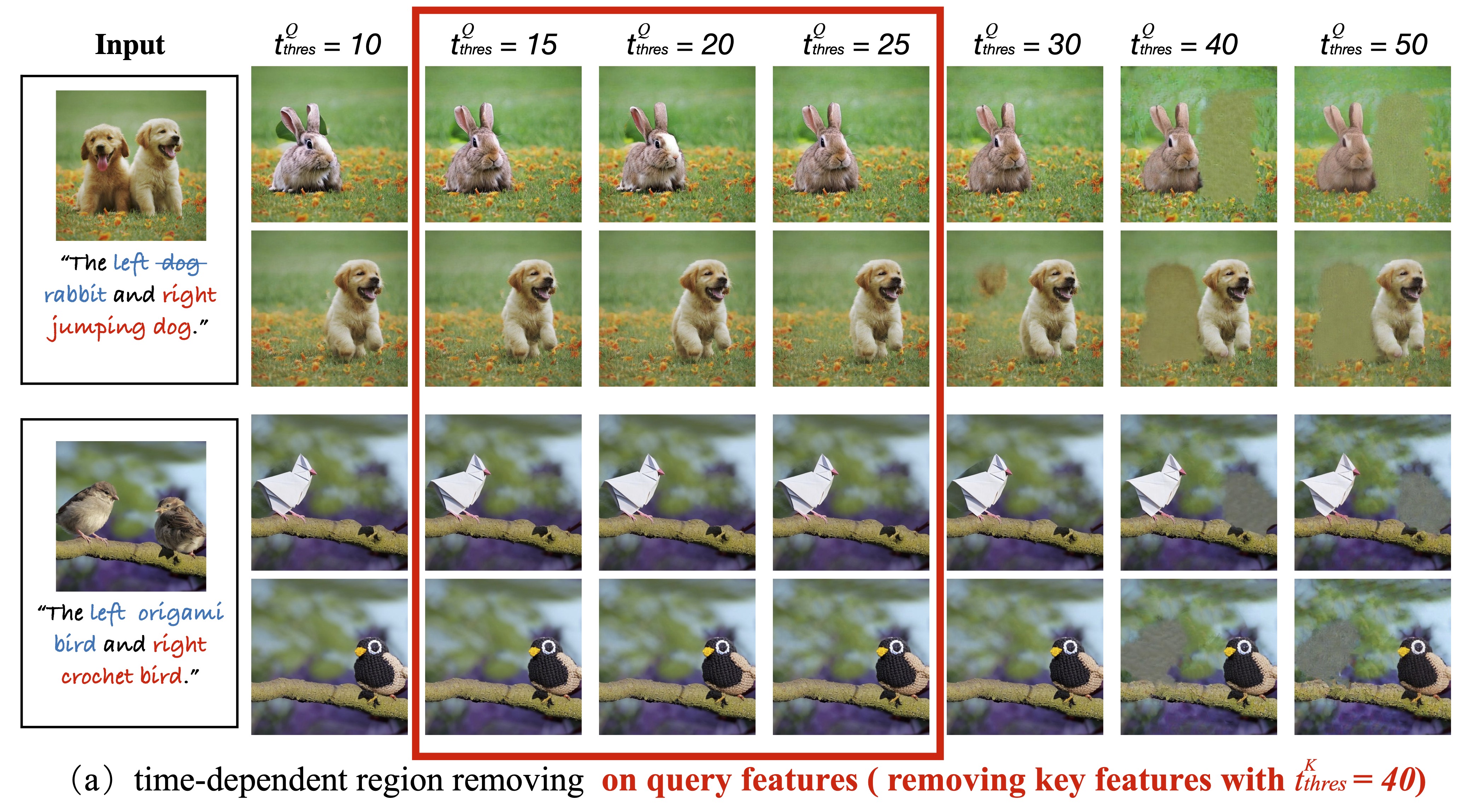} 
\caption{Ablation results on parameter $t^Q_{thres}$ with time-dependent region removing scheme acting  on self-attention's query $\mathcal{Q}$ and key features ($t^K_{thres}=40$).
} 
\label{QK}
\end{figure*}

The model adopts the control variable method to visualize the results of applying the time-dependent region removing scheme to query $\mathcal{Q}$, key $\mathcal{K}$, and value $\mathcal{V}$ features in self-attention, respectively. 
As shown in Fig.\textcolor{blue}{\ref{QKV-ablation}}: (1) the region feature removing operation on value features inherently leads to inconsistent shadows, as the value features are not normalized by the softmax operation, rendering them more sensitive to outliers post-removal..
(2) The region feature removing operation on query features, significantly disrupts the object's conflicting region, enhancing object-layered editability. However, feature removing based solely on the query features lacks robustness, especially when query features are severely disrupted, so that a large $t_thres^Q$ may hinder the harmonious complementation of the removed region.
(3) The region feature removing operation on key feature balances both the removal of the conflict regions and the harmonious complementation of removed regions. 
The final setting combines the removal of query and key features for the joint advancements of object-layered editability and coherent complementation of removed regions, as shown in Fig.\textcolor{blue}{\ref{QK}}. 
Within a reasonable range (red box marks, $t^K_{thres}= \in [35, 45], t^Q_{thres}= \in [15, 25]$), the model effectively removes inter-object conflicting regions while achieving harmonious completion, confirming the effectiveness and robustness of our conflict-aware multi-layer learning model.





\begin{table}
  \centering
    \resizebox{1\linewidth}{!}{
 \begin{tabular}{lcccc}
    \toprule
    &  \multicolumn{2}{c}{\multirow{1}{*}{\textbf{Multi-aspect}}} &  \multicolumn{2}{c}{\multirow{1}{*}{\textbf{Multi-object}}}  \\ 
  &{\textbf{{Text-Align}}}      &{\textbf{{Image-Align}}}   & {\textbf{{Text-Align}}}      &{\textbf{{Image-Align}}}  \\
    \midrule
   OIR &17.1\% &12.5\% &10.1\% &7.6\%  \\
     GenArtist &7.9\%  & 12.7\%   &6.2\% &8.4\% \\
    ParalllelEdits &19.6\% & 8.3\%  &8.1\% &9.7\%   \\
    LoMOE &13.7\%  &20.9\%  &13.7\% &14.6\%\\
    h-Edit&10.4\%  &16.5\% &9.5\% &4.9\%\\
    \textbf{Ours}(SDXL) &\textbf{31.3\%} &\textbf{28.1\%} &\textbf{52.4\%} &\textbf{54.8\%} \\
    \bottomrule
  \end{tabular}}
  \caption{{Human Evaluation}. \textit{LayerEdit} outperforms baselines in all metrics.}
  \label{human-T}
\end{table}


\section{Human Evaluation }

For multi-aspect and multi-object editing, we conduct a human preference study to compare our \textit{LayerEdit} with five SOTA baselines. $10$ participants are required to evaluate $100$ cases ($50$ each for multi-aspect and multi-object editing) and $1200$ generated images from two aspects: (1) text-alignment: measuring the consistency with edited text; (2) image-alignment: measuring the consistency with the text-irrelevant image details.     
As indicated in Tab.\textcolor{blue}{\ref{human-T}}, our method outperforms existing methods, especially for multi-function editing, with a preference rate of more than $50\%$ across all baselines.


\end{document}